# Learning Bayesian Network Equivalence Classes with Ant Colony Optimization

**Rónán Daly**                              RDALY@DCS.GLA.AC.UK
*Department of Computing Science*
*University of Glasgow*
*Sir Alwyn Williams Building*
*Glasgow, G12 8QQ, UK*

**Qiang Shen**                                QQS@ABER.AC.UK
*Department of Computer Science*
*Aberystwyth University*
*Penglais Campus*
*Aberystwyth, SY23 3DB, UK*

## Abstract

Bayesian networks are a useful tool in the representation of uncertain knowledge. This paper proposes a new algorithm called ACO-E, to learn the structure of a Bayesian network. It does this by conducting a search through the space of equivalence classes of Bayesian networks using Ant Colony Optimization (ACO). To this end, two novel extensions of traditional ACO techniques are proposed and implemented. Firstly, multiple *types* of moves are allowed. Secondly, moves can be given in terms of indices that are not based on construction graph nodes. The results of testing show that ACO-E performs better than a greedy search and other state-of-the-art and metaheuristic algorithms whilst searching in the space of equivalence classes.

## 1. Introduction

The task of learning Bayesian networks from data has, in a relatively short amount of time, become a mainstream application in the process of knowledge discovery and model building (Aitken, Jirapech-Umpai, & Daly, 2005; Heckerman, Mamdani, & Wellman, 1995). The reasons for this are many. For one, the model built by the process has an intuitive feel – this is because a Bayesian network consists of a directed acyclic graph (DAG), with conditional probability tables annotating each node. Each node in the graph represents a variable of interest in the problem domain and the arcs can (with some caveats) be seen to represent causal relations between these variables (Heckerman, Meek, & Cooper, 1999) – the nature of these causal relations is governed by conditional probability tables associated with each node/variable. An example Bayesian network is shown in Figure 1.

Another reason for the popularity of Bayesian networks is that aside from the visual attractiveness of the model, the underlying theory is quite well understood and has a solid foundation. A Bayesian network can be seen as a factorization of a joint probability distribution, with the conditional probability distributions at each node making up the factors and the graph structure making up their method of combination. Because of this equivalence, the network can answer any probabilistic question regarding the variables modeled.

In addition, the popularity of Bayesian networks has been increased by the accessibility of methods to query the model and learn both the structure and parameters of the network (Daly, Shen, & Aitken, 2009). It has been shown that inference in Bayesian networks is NP-complete (Dagum & Luby, 1993; Shimony, 1994), but approximate methods have been found to perform this operation in an acceptable amount of time. Learning the structure of Bayesian networks is also NP-complete (Chickering, 1996a), but here too, methods have been found to render this operation tractable. These include greedy search, iterated hill climbing and simulated annealing (Chickering, Geiger, & Heckerman, 1996). Recently however, other heuristics have become popular with the problem of combinatorial optimization in high dimensional spaces. These include approaches such as tabu search (Glover, 1989, 1990), genetic algorithms (Mitchell, 1996) and – the approach that this paper will investigate – Ant Colony Optimization (ACO).

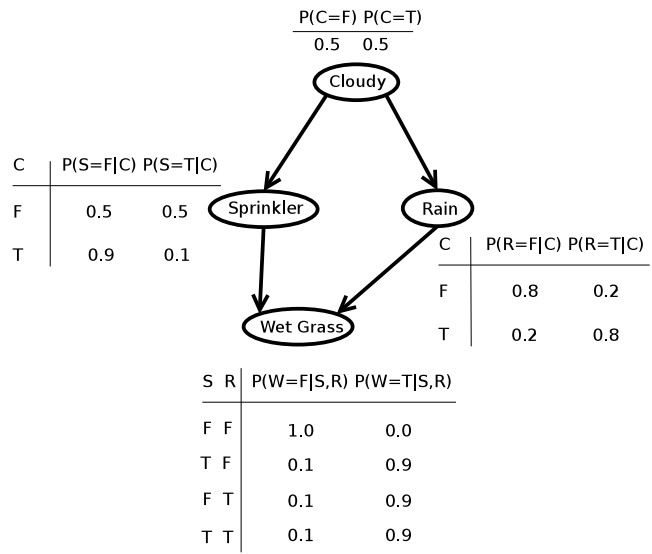

Figure 1: An example Bayesian network

ACO is a fairly recent, so called metaheuristic, that is used in the solution of combinatorially hard problems (Dorigo & Stützle, 2004). It is an iterated, stochastic technique that is biased by the results of previous iterations (Birattari, Caro, & Dorigo, 2002). The method is modeled on the behavior of real-life ants foraging for food.

Many ants secrete a pheromone trail that is recognizable by other ants and which positively biases them to follow that trail, with a stronger trail meaning it is more likely to be biased towards it. Over time this pheromone trail evaporates. When hunting for food, an ant's behavior is to randomly walk about, perhaps by following a pheromone trail, until it finds some food. It then returns in the direction from whence it came. Because the strength of the trail is a factor in choosing to follow it, if an ant is faced with two or more pheromone trails to choose from, it will tend to choose the trails with the highest concentration of pheromone.

With these characteristics, in a situation where there are multiple paths to a food source, ants generally follow the shortest path. This can be explained as follows. Assuming ants start from a nest and no pheromone trails are present, they will randomly wander until they reach a food source and then return home, laying pheromone on the way back. The ant that chooses the shortest path to the food source will return home the quickest, which means their pheromone trail will have the highest concentration, as more pheromone is laid per unit of time. This stronger trail will cause other ants to prefer it over longer trails. These ants will then leave their own pheromone on this short trail, thereby providing a reinforcing behavior to choose this trail over others.

As a computing technique, ACO is roughly modeled on this behavior. Artificial ants walk around a graph where the nodes represent pieces of a solution. They continue this until a complete solution is found. At each node, a choice of the next edge to traverse is made, depending on a pheromone value associated with the edge and a problem specific heuristic. After a number of ants have performed a traversal of the graph, one of the best solutions is chosen and the pheromone on the edges that were taken is increased, relative to the other edges. This biases the ants towards choosing these edges in future iterations. The search stops when a problem specific criterion is reached. This could be stagnation in the quality of solutions or the passage of a fixed amount of time.

This paper will seek to use the ACO technique in learning Bayesian networks. Specifically, it will be used to learn an equivalence class of Bayesian network structures. To this end, the rest of this paper will be structured in the following fashion. Firstly, there will be a more in-depth study of the problem of searching for an optimum Bayesian network, in both the space of Bayesian networks themselves and of equivalence classes of Bayesian networks. Then, a new method of formulating a search for a Bayesian network structure in terms of the ACO metaheuristic will be introduced. This method is based in part on earlier work done on this topic (Chickering, 2002a; de Campos, Fernández-Luna, Gámez, & Puerta, 2002). Next, results of tests against previous techniques will be discussed and finally, conclusions and possible future directions will be stated.

## 2. Searching for a Bayesian Network Structure

There are, in general, three different methods used in learning the structure of a Bayesian network from data. The first finds conditional independencies in the data and then uses these conditional independencies to produce the structure (Spirtes, Glymour, & Scheines, 2000). Probably the most well known algorithms that use this method are the PC algorithm by Spirtes and Glymour (1990) and the CI and FCI algorithms of Spirtes, Meek, and Richardson (1995) that are able to identify latent variables and selection bias. The second uses dynamic programming and optionally, clustering, to construct a DAG (Ott, Imoto, & Miyano, 2004; Ott & Miyano, 2003). The third method – which is to be dealt with here – defines a search on the space of Bayesian networks. This method uses a scoring function defined by the implementer, which says relatively how good a network is compared to others.

Although the classification into three different methods as noted above is useful in differentiating their applicability, the boundaries between them are often not as clear as they may seem. E.g. the score and search approach and the dynamic programming approach are both similar in that they use scoring functions. Indeed, there is a view by Cowell (2001) that the conditional independence approach is equivalent to minimizing the Kullback-Leibler (KL) divergence (Kullback & Leibler, 1951) using the score and search approach. Before discussing how the score and search method works, some definitions and notation will be introduced.

A graph $\mathcal{G}$ is given as a pair $(V, E)$, where $V = \{v_1, \ldots, v_n\}$ is the set of vertices or nodes in the graph and $E$ is the set of edges or arcs between the nodes in $V$. A directed graph is a graph where all the edges have an associated direction from one node to another. A directed acyclic graph or DAG, is a directed graph without any cycles, i.e. it is not possible to return to a node in the graph by following the direction of the arcs. For illustration, the graph in Figure 2 is a DAG. The parents of a node $v_i$, $Pa(v_i)$, are all the nodes $v_j$ such that there is an arrow from $v_j$ to $v_i$ ($v_j \rightarrow v_i$). The descendants of $v_i$, $D(v_i)$, are all the nodes (not including $v_i$) reachable from $v_i$ by following the arrows in a forwards direction repeatedly. The non-descendants of $v_i$, $ND(v_i)$, are all the nodes (not including $v_i$) that are not descendants of $v_i$.

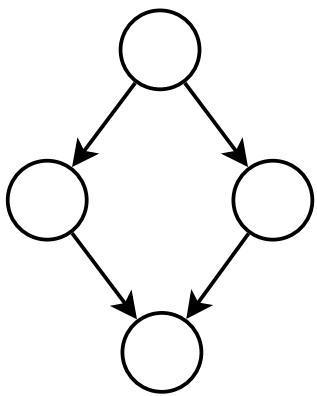 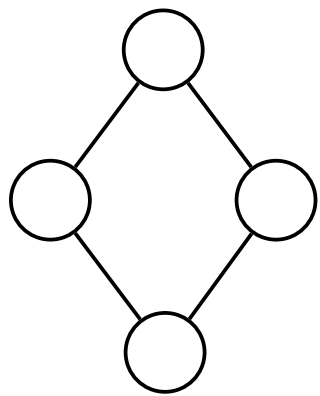

Figure 2: A directed acyclic graph          Figure 3: The skeleton of the DAG in Figure 2

Let there be a graph $\mathcal{G} = (V, E)$ and a joint probability distribution $P$ over the nodes in $V$. Let $I_P(X, Y | Z)$ mean that each of the variables in set $X$ is conditionally independent of each of the variables in set $Y$ under probability distribution $P$ given the variables in set Z. Say also that the following is true

$$\forall v \in V. I_P(\{v\}, ND(v) | Pa(v)).$$

That is, each node is conditionally independent of its non-descendants, given its parents. Then it is said that $\mathcal{G}$ satisfies the Markov condition with $P$, and that $(\mathcal{G}, P)$ is a Bayesian network. Notice the conditional independencies implied by the Markov condition. They allow the joint distribution $P$ to be written as the product of conditional distributions; $P(v_1 | Pa(v_1)) P(v_2 | Pa(v_2)) \cdots P(v_n | Pa(v_n)) = P(v_1, v_2, \ldots, v_n)$. However, more importantly, the reverse can also be true. Given a DAG $\mathcal{G}$ and either discrete conditional distributions or certain types of continuous conditional distributions (e.g. Gaussians), of the form $P(v_i | Pa(v_i))$ then there exists a joint probability distribution

$$P(v_1, v_2, \ldots, v_n) = P(v_1 | Pa(v_1)) P(v_2 | Pa(v_2)) \cdots P(v_n | Pa(v_n)).$$

This means that if we specify a DAG – known as the structure – and conditional probability distributions for each node given its parents, which are often parameterised, we have a Bayesian network, which is a representation of a joint probability distribution.

In learning a Bayesian network from data, both the structure $\mathcal{G}$ and parameters of the conditional probability distributions $\Theta$ must be learned, normally separately. In the case of complete multinomial data, the problem of learning the parameters is easy given certain reasonable assumptions, with a simple closed form formula for $\Theta$ (Heckerman, 1995). However, in the case of learning the structure, no such formula exists and other methods are needed. In fact, learning the optimal structure with discrete variables is an NP-hard problem in almost all circumstances and consequently enumeration and test of all network structures is not likely to succeed (Chickering, 1996a; Chickering, Heckerman, & Meek, 2004). With just ten variables there are roughly $10^{18}$ possible DAGs. Whilst there exist dynamic programming methods that can handle roughly 30 variables as discussed above, in general, non-exact and heuristic methods are possibly the only tractable solution to anything above this.

In order to create a space in which to search through, three components are needed. Firstly all the possible solutions must be identified as the set of states in the space. Secondly a representation mechanism for each state is needed. Finally a set of operators must be given, in order to move from state to state in the space.

Once the search space has been defined, two other pieces are needed to complete the search algorithm, a scoring function which evaluates the "goodness of fit" of a structure with a set of data and a search procedure that decides which operator to apply, normally using the scoring function to see how good a particular operator application might be. An example of a search procedure is greedy search, which at every stage applies the operator that produces the best change in the structure, according to the scoring function. As for the scoring function, various formulæ have been found to see how well a DAG matches a data sample.

One of these functions is given by computing the relative posterior probability of a structure $\mathcal{G}$ given a sample of data $D$, i.e.

$$S(\mathcal{G}, D) = P(\mathcal{G}, D) = P(D|\mathcal{G})P(\mathcal{G}).$$

The likelihood term above can take many forms. One popular method is called the Bayesian Dirichlet (BD) metric. Here,

$$P(D|\mathcal{G}) = \prod_{i=1}^{n} \prod_{j=1}^{q_i} \frac{\Gamma(N'_{ij})}{\Gamma(N'_{ij} + N_{ij})} \cdot \prod_{k=1}^{r_i} \frac{\Gamma(N'_{ijk} + N_{ijk})}{\Gamma(N'_{ijk})} \tag{1}$$

In this formula, there are $n$ variables in the graph, so the first product is over each variable. There are $q_i$ configurations of the parents of node $i$, so the second product is over all possible parent configurations, i.e. the Cartesian product of the number of possible values each parent variable can take. Each variable $i$ can take on one of $r_i$ possible values. The value $N_{ijk}$ is the number of times that variable $i = k$ and the parents of $i$ are in configuration $j$ in the data sample $D$. $N_{ij}$ is given as $\sum_{k=1}^{r_i} N_{ijk}$, i.e. the sum of $N_{ijk}$ over all possible values that $i$ can take on. With $N'_{ij} = \sum_{i=1}^{r_i} N'_{ijk}$, the values $N'_{ijk}$ are given as parameters that give different variants of the BD metric. E.g. if $N'_{ijk}$ is set to 1 the K2 metric results, as given by Cooper and Herskovits (1992). With $N'_{ijk}$ set to $N'/(r_i \cdot q_i)$ (where $N'$, known as the equivalent sample size is a measure of the confidence in the prior network), the BDeu metric results which was proposed by Buntine (1991) and further generalised by Heckerman, Geiger, and Chickering (1995).

The prior value $P(\mathcal{G})$ is a measure of how probable a particular structure is before any data is seen. These values can often be hard to estimate because of the massive numbers of graphs, each of them needing a probability. Therefore, the values are often given as uniform over all possible network structures or possibly favouring structures with less arcs.

Other forms used for the scoring function are $S(\mathcal{G}, D) = \log P(D|\mathcal{G}, \hat{\Theta}) - \frac{d}{2} \log N$, known as the Bayesian information criterion (BIC) (Schwarz, 1978) and $S(\mathcal{G}, D) = \log P(D|\mathcal{G}, \hat{\Theta}) - d$, known as the Akaike Information Criterion (AIC) (Akaike, 1974). In these models, the parameters $\hat{\Theta}$ give the maximum likelihood estimate of the likelihood, $d$ is the number of free parameters in the structure and $N$ is the number of samples in the data $D$.

Traditionally, in searching for a Bayesian network structure, the set of states is the set of possible Bayesian network structures, the representation is a DAG and the set of operators are various small local changes to a DAG, e.g. adding, removing or reversing an arc, as illustrated in Table 1. This type

| Operator | Before | | After | |
|---|---|---|---|---|
| Insert_Arc(X,Y) | 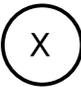 | 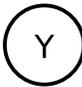 | 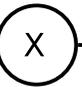 | 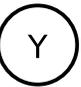 |
| Delete_Arc(X,Y) | 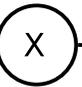 | 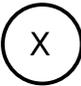 | 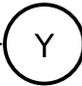 | 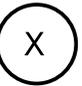 |
| Reverse_Arc(X,Y) | 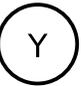 | 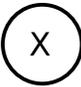 | 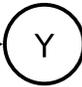 | 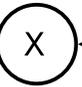 |

Table 1: Basic modification operators

of search is convenient because of the decomposition properties of score functions,

$$S(\mathcal{G}, D) = \prod_{i=1}^{n} s\left(v_i, \mathrm{Pa}^{\mathcal{G}}(v_i), D\right),$$

where $s$ is a scoring function that takes a node $v_i$ and the parents of this node in graph $\mathcal{G}$, $\mathrm{Pa}^{\mathcal{G}}(v_i)$. Popular scoring functions such as the BD metric are decomposable in this manner. Successful application of the operators is also dependent on the changed graph being a DAG, i.e. that no cycle is formed in applying the operator.

## 3. Searching in the Space of Equivalence Classes

According to many scoring criteria, there are DAGs that are equivalent to one another, in the sense that they will produce the same score as each other. It has been known for some time that these DAGs are equivalent to one another, in that they entail the same set of independence constraints as each other, even though the structures are different. According to a theorem by Verma and Pearl (1991), two DAGs are equivalent if and only if they have the same skeletons and the same set of v-structures. The 'skeleton' is the undirected graph that results in undirecting all edges in a DAG (see Figure 3) and a 'v-structure' (sometimes referred to as a morality), is a head-to-head meeting of two arcs, where the tails of the arcs are not joined. These concepts are illustrated in Figure 4. From this notion of equivalence, a class of DAGs that are equivalent to each other can be defined, notated here as $Class(\mathcal{G})$.

### 3.1 Representation of Equivalence Classes

Because of this apparent redundancy in the space of DAGs, attempts have been made to conduct the search for Bayesian network structures in the space of equivalence classes of DAGs (Acid & de Campos, 2003; Chickering, 1996b, 2002a; Munteanu & Bendou, 2001). The search set of this space is the set of equivalence classes of DAGs and will be referred to as E-space. To represent the members of this equivalence class, a different type of structure is used, known as a partially directed acyclic graph (PDAG). A PDAG (an example of which is shown in Figure 5) is a graph that may contain both undirected and directed edges and that contains no directed cycles and will be notated

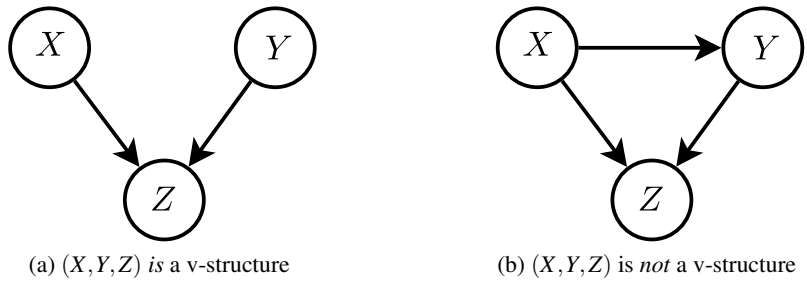

(a) $(X, Y, Z)$ *is* a v-structure            (b) $(X, Y, Z)$ is *not* a v-structure

Figure 4: V-Structures

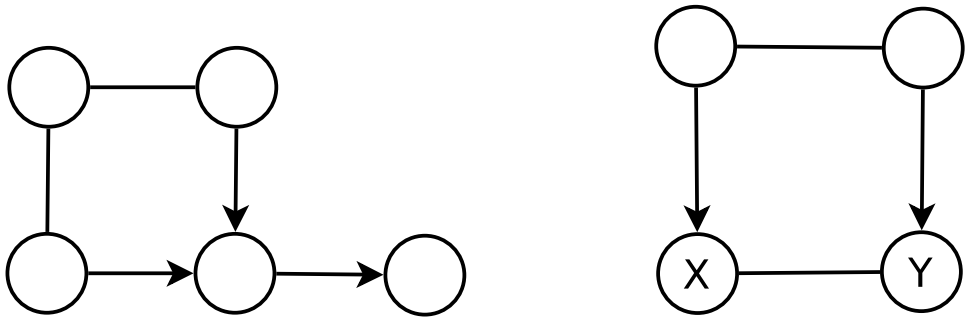

Figure 5: A partially directed acyclic graph

Figure 6: A PDAG for which there exists no consistent extension

herein as $\mathcal{P}$. The equivalence class of DAGs corresponding to a PDAG is denoted as $Class(\mathcal{P})$, with a DAG $\mathcal{G} \in Class(\mathcal{P})$ if and only if $\mathcal{G}$ and $\mathcal{P}$ have the same skeleton and same set of v-structures.

Related to this is the idea of a *consistent extension*. If a DAG $\mathcal{G}$ has the same skeleton and the same set of v-structures as a PDAG $\mathcal{P}$ then it is said that $\mathcal{G}$ is a consistent extension of $\mathcal{P}$. Not all PDAGs have a DAG that is a consistent extension of itself. If a consistent extension exists, then it is said that the PDAG *admits* a consistent extension. Only PDAGs that admit a consistent extension can be used to represent an equivalence class of DAGs and hence a Bayesian network. An example of a PDAG that does not have a consistent extension is shown in Figure 6. In this figure, directing the edge $x - y$ either way will create a v-structure that does not exist in the PDAG and hence no consistent extension can exist.

Directed edges in a PDAG can be either: compelled, or made to be directed that way; or reversible, in that they could be undirected and the PDAG would still represent the same equivalence class. From this idea, a completed PDAG (CPDAG) can be defined, where every undirected edge is reversible in the equivalence class and every directed edge is compelled in the equivalence class. Such a CPDAG will be denoted as as $\mathcal{P}^{\mathcal{C}}$. It can be shown that there is a one-to-one mapping between a CPDAG $\mathcal{P}^{\mathcal{C}}$ and $Class(\mathcal{P}^{\mathcal{C}})$. Therefore, by supplying a CPDAG, one can uniquely denote a set of conditional independencies. For a more in-depth look at this topic, see the papers of Andersson, Madigan, and Perlman (1997) and Chickering (1995).

## 3.2 Techniques for Searching through Equivalence Classes

Note that below, a *move* is referred to as an application of an operator to a particular state in the search space.

To be able to conduct a search through the space of equivalence classes, a method must be able to find out whether a particular move is valid and if valid, how good that move is. These tasks are relatively easy whilst searching through the space of DAGs – a check whether a move is valid is equivalent to a check whether a move keeps a DAG acyclic. The goodness of such a move is found out by using the scoring function, but rather than scoring each neighboring DAG in the search space, the decomposability of most scoring criteria can be taken advantage of, with the result that only nodes whose parent sets have changed need to be scored.

However, this task of checking move validity and move score is not as easy in the space of equivalence classes. These classes are often represented by PDAGs, as discussed in the previous section. For one, instead of just checking for cycles, checks also have to be made so that unintended v-structures are not created in a consistent extension of a PDAG. Scoring a move also creates difficulties, as it is hard to know what extension and hence what changes in parent sets of nodes will occur, without actually performing this extension. Also, a local change in a PDAG *might* make a non-local change in a corresponding consistent extension and so force unnecessary applications of the score function.

These problems were voiced as concerns by Chickering (1996b). In that paper, validity checking of moves is performed by trying to obtain a consistent extension of the resulting PDAG – if none exists then the move is not valid. Scoring the move was achieved by scoring the changed nodes in the consistent extension given. These methods were very generic, but resulted in a significant slowdown in algorithm execution, compared to search in the space of DAGs.

To alleviate this problem, authors proposed improvements that would allow move validity and move score to be computed without needing to obtain a consistent extension of the PDAG (Acid & de Campos, 2003; Chickering, 2002a; Munteanu & Bendou, 2001). This was done by defining an explicit set of operators, with each operator having a validity test and corresponding score change function, that could be calculated on the PDAG. These changes led to a speedup of the execution time of the algorithm, with the result that search in the space of equivalence classes of Bayesian networks became competitive with search in the space of Bayesian networks. An example of one set of these operators is given in Table 2. In this table, the variables $x$ and $y$ refer to nodes in a graph. As an example, the InsertU operator takes two nodes as arguments, $x$ and $y$. It can be seen that all the operators take two arguments, except MakeV, which takes three arguments. Each operator also has a set of validity tests that must be passed in order for the application of the operator with its particular arguments to be valid. Finally, the score difference between the old and new PDAGs is given in the last column.

Note that in this table:

$\Pi_x$ is the parent set of node $x$, i.e. the set of nodes that have directed arcs going to node $x$;

$N_x$ is the neighbor set of node $x$, i.e. the set of nodes that have undirected arcs going to node $x$;

$N_{x,y}$ is the set of shared neighbors of nodes $x$ and $y$, i.e. $N_x \cap N_y$; and

$\Omega_{x,y}$ is the set of parents of $x$ that are neighbors of $y$, i.e. $\Pi_x \cap N_y$.

| Operator | Effect | Validity Tests | Change in Score |
|---|---|---|---|
| InsertU<br>$x - y$ | Add an undirected arc between $x$ and $y$ | 1. Every undirected path from $x$ to $y$ contains a node in $N_{x,y}$<br>2. $\Pi_x = \Pi_y$ | $s\left(y, N_{x,y}^{+x} \cup \Pi_y\right)$<br>$- s\left(y, N_{x,y} \cup \Pi_y\right)$ |
| DeleteU<br>$x - y$ | Delete an undirected arc between $x$ and $y$ | $N_{x,y}$ is a clique | $s\left(y, N_{x,y} \cup \Pi_y\right)$<br>$- s\left(y, N_{x,y}^{+x} \cup \Pi_y\right)$ |
| InsertD<br>$x \rightarrow y$ | Add a directed arc from $x$ to $y$ | 1. Every semi-directed path from $y$ to $x$ contains a node in $\Omega_{x,y}$<br>2. $\Omega_{x,y}$ is a clique<br>3. $\Pi_x \neq \Pi_y$ | $s\left(y, \Omega_{x,y} \cup \Pi_y^{+x}\right)$<br>$- s\left(y, \Omega_{x,y} \cup \Pi_y\right)$ |
| DeleteD<br>$x \rightarrow y$ | Delete a directed arc from $x$ to $y$ | $N_y$ is a clique | $s\left(y, N_y \cup \Pi_y^{-x}\right)$<br>$- s\left(y, N_y \cup \Pi_y\right)$ |
| ReverseD<br>$x \rightarrow y$ | Reverse a directed arc from $x$ to $y$ | 1. Every semi-directed path from $x$ to $y$ that does not include the edge $x \rightarrow y$ contains a node in $\Omega_{y,x} \cup N_y$<br>2. $\Omega_{y,x}$ is a clique | $s\left(y, \Pi_y^{-x}\right)$<br>$+ s\left(x, \Pi_x^{+y} \cup \Omega_{y,x}\right)$<br>$- s\left(y, \Pi_y\right)$<br>$- s\left(x, \Pi_x \cup \Omega_{y,x}\right)$ |
| MakeV<br>$x \rightarrow z \leftarrow y$ | Direct undirected arcs from $x$ and $y$ to $z$ | Every undirected path between $x$ and $y$ contains a node in $N_{x,y}$ | $s\left(z, \Pi_z^{+y} \cup N_{x,y}^{-z+x}\right)$<br>$+ s\left(y, \Pi_y \cup N_{x,y}^{-z}\right)$<br>$- s\left(z, \Pi_z \cup N_{x,y}^{-z+x}\right)$<br>$- s\left(y, \Pi_y \cup N_{x,y}\right)$ |

Table 2: Validity conditions and change in score for each operator

Also, as a convenience, $M^{+x}$ is notation for $M \cup \{x\}$ and $M^{-x}$ is notation for $M \setminus \{x\}$.

This notation and the set of operators in Table 2 come from those proposed by Chickering (2002a). Other definitions include: an undirected path is a path from one node to another that only follows undirected edges; a semi-directed path is a path from one node to another that only follows undirected edges or directed edges from tail to head; and a set of nodes $N$ is a clique, if it is a completely connected subgraph of a graph, (i.e. every node is connected to every other in the subgraph).

### 3.3 Advantages of Searching in E-space

With this representation of equivalence classes of Bayesian network structures and a set of operators that modify the CPDAGs which represent them (e.g. insert an undirected arc, insert a directed arc etc.), a search procedure can proceed. However, what reasons are there for pursuing this type of search? Chickering (2002a) gives a list of reasons, some of which are discussed here.

For one, an equivalence class can represent many different DAGs in a single structure. With a DAG representation, time can be wasted rescoring DAGs that are in the same equivalence class. And with a search in the space of DAGs, the connectivity of the search space can mean that the ability to move to a particular neighboring equivalence class can be constrained by the particular representation given by a DAG. There is also the problem given by the prior probability used in the scoring function. Whilst searching through the space of DAGs, certain equivalence classes can be over represented by this prior, because there are many more DAGs contained in the class. An example can be given in the case of networks with two nodes. In B-space there are 3 possible structures, which with equal priors give $P(\mathcal{G}) = 1/3$, for each DAG $\mathcal{G}$. However, the two DAGs that are connected represent the same equivalence class, giving it an effective prior of $2/3$. In E-space there are 2 possible structures, which with equal priors give $P(\mathcal{P}) = 1/2$, for each PDAG $\mathcal{P}$. This is not necessarily a problem when performing model *selection*, but becomes much more of an issue when performing model *averaging*.

These concerns have motivated researchers. In particular, recent implementations of algorithms that search through the space of equivalence classes have produced results that show a marked improvement in execution time and a small improvement in learning accuracy, depending on the type of data set (Chickering, 2002a,b).

## 4. Ant Colony Optimization

Ant colony optimization is a global optimization technique generally used in the area of combinatorial problems, i.e. problems where the set of solutions is discrete. Since the inception of its present form by Dorigo (1992), ACO has been successfully applied to many combinatorially hard problems including the sequential ordering problem (Gambardella & Dorigo, 2000), the vehicle routing problem (Bullnheimer, Hartl, & Strauss, 1999), the bin-packing problem (Levine & Ducatelle, 2004) and many more (Costa & Hertz, 1997; Gambardella & Dorigo, 2000; Maniezzo & Colorni, 1999; Stützle, 1998). Such a diverse range of applications must ask the question as to what is the nature of the system that can solve them.

The particular form of ACO is of a *metaheuristic* in the field of *swarm intelligence* (Bonabeau, Dorigo, & Theraulaz, 1999), that is based on the behavior of real-life ants as they forage for food. A metaheuristic is a general purpose heuristic that guides other, more problem specific heuristics, whilst swarm intelligence may be defined as:

> 'algorithms or distributed problem-solving devices inspired by the collective be-
> haviour of social insect colonies and other animal societies' (Bonabeau, Dorigo et al.,
> 1999).

It is in this conceptual framework that ACO is defined.

## 4.1 Ant Colony Optimization

Ant colony optimization is a swarm intelligence technique that is based on the foraging behavior of real-life ants. In particular, it uses the principle of stigmergy (the indirect communication of agents through the environment) as a communication mechanism. Real-life ants leave a chemical trail behind them as they explore their environment. This trail is known as *pheromone*. In moving around, ants are more likely to follow a path with more pheromone, than a path with less (or no) pheromone. This behavior was investigated by Deneubourg, Aron, Goss, and Pasteels (1990), who designed an experiment with a nest of Argentine ants, a food source and two trails between them that could be set to different length. Ants would leave the nest, find the food source and return back with food. When the trails were of the same length, it was found that the ants would eventually settle on a single trail for travel to and from the nest. This behavior can be explained as follows.

When the experiment begins, ants initially choose one of the trails at random. Whilst traversing this trail, they deposit pheromone. This causes following ants to choose the trail the initial ants took more often, and deposit more pheromone on that trail. Again, this causes more ants to choose the initially chosen trail, *to a greater degree than the first set of ants*. Put another way, each ant that chooses a certain trail reinforces the probability that following ants will choose that trail. The trail that initially gets chosen by more ants has more pheromone deposited per unit time and hence a positive feedback or *autocatalytic* process is created, where eventually all ants converge to a single trail.

When the trails start out at different lengths, it is found that ants converge on the shorter trail more often than the longer. This can be explained by more ants being able to traverse the shorter trail to the food source and return to the nest in the same amount of time it would take to traverse the longer trail. With more ants traversing the trail, more pheromone is deposited, and the ants eventually converge to that path.

It is the behavior of the ants when faced with trails of different lengths that ACO is modeled upon. Instead of real-life ants, artificial ants are conceived as a computing unit. Instead of trails, these ants traverse a construction graph. The paths the ants take on this graph are solutions to the problem being looked at – the idea is to reinforce the pheromone on better solutions. However, the fundamental idea of laying down pheromone is kept, with ants depositing it on arcs as they traverse from node to node. Also, ants are programmed to follow arcs with stronger pheromone more often than arcs with weaker pheromone.

Artificial ants can be more useful than real-life ants in that they can be given a memory. This can stop ants looping around and helps when laying pheromone on the return journey. Also they can be programmed to use problem dependent heuristics, which can guide the search towards better solutions. All of these ideas and more will now be discussed.

## 4.2 The ACO Metaheuristic

Nowadays, ACO algorithms tend to be defined in terms of the ACO metaheuristic (Dorigo & Di Caro, 1999). A metaheuristic is a general purpose heuristic that guides other, more problem specific heuristics. Examples of metaheuristics include simulated annealing (Kirkpatrick, Gelatt, & Vecchi, 1983), tabu search (Glover, 1989, 1990), evolutionary computation etc.

In the ACO metaheuristic, a problem is represented by a triple $(\mathcal{S}, f, \Omega)$, where $\mathcal{S}$ is a set of candidate solutions, $f : \mathcal{S} \times T$ is an objective or scoring function that measures a solution's quality at a particular time $t \in T$ and $\Omega : T$ is a set of constraints at time $t \in T$, used in a solution's construction. The range of $f$ and $\Omega$ is dependent on the particular instance of the metaheuristic. In trying to map a combinatorial optimization problem onto this representation, the following framework is used.

- There should be a finite set of solution components $C = \{c_1, c_2, \ldots, c_{N_c}\}$. These are the building blocks of candidate solutions.

- The problem states are represented by sequences of solution components $x = \langle c_i, c_j, \ldots \rangle$. The set of all possible sequences is given as $\mathcal{X}$.

- $\mathcal{S}$ – the set of candidate solutions as mentioned above – is a subset of $\mathcal{X}$, i.e. $\mathcal{S} \subseteq \mathcal{X}$.

- There is a set of feasible states $\tilde{\mathcal{X}}$, with $\tilde{\mathcal{X}} \subseteq \mathcal{X}$. A feasible state $x \in \tilde{\mathcal{X}}$ is a state where it is possible to add components from $C$ to $x$ to create a solution satisfying the constraints $\Omega$.

- Each candidate solution $s \in \mathcal{S}$ has a cost $g(s,t)$. Normally $g(s,t) \equiv f(s,t)$, $\forall s \in \tilde{\mathcal{S}}$, where $\tilde{\mathcal{S}} = \mathcal{S} \cap \tilde{\mathcal{X}}$ is the set of feasible candidate solutions. However, this might not always be the case; if $f$ is very expensive to compute, $g$ might be an easier to compute function that is broadly similar to $f$ and that can be used in the generation of solutions.

- The set of optimal solutions $\mathcal{S}^*$ should be non-empty, with $\mathcal{S}^* \subseteq \tilde{\mathcal{S}}$.

- Sometimes it may also be possible to associate a cost $J(x,t)$ to a state $x \in \mathcal{X}$ that is not a candidate solution.

With this framework, solutions to the problem $(S, f, \Omega)$ can be generated by having artificial ants perform a *random walk* on the complete graph $G$ defined on the components in $C$. This graph $G$ is known as the *construction graph*. A random walk on a graph is a series of moves from node to node of the graph, with each move being random to some degree. If the walk is *Markovian*, then the next move is always completely random; if not then then next move is influenced by the previous moves. Hence, using this terminology ACO is non-Markovian. The walk that the ant makes is generally biased by two things – a heuristic value $\eta$ ($\eta_i$ if the heuristic is associated with the individual nodes of $G$, $\eta_{ij}$ if it is associated with the edges of $G$) and a pheromone trail $\tau$ (again, $\tau_i$ if the pheromone is associated with the individual nodes of $G$, $\tau_{ij}$ if the pheromone is associated with the edges of $G$). The way the heuristic and pheromone are implemented are problem dependent, but in general the heuristic $\eta$ is a measure of the 'goodness' of taking a particular move on the construction graph as defined by some local measure. The pheromone $\tau$ is a measure of the 'goodness' of taking a particular move as defined by the aggregate behavior of ants selecting that move and the quality of solutions that these ants generate.

Finally, each artificial ant $k$ has the following properties in order to fully specify how the random walk will proceed:

**Memory –** Each ant $k$ has a memory $\mathcal{M}^k$ that stores information about the path it has so far followed.

**Start State –** Each ant $k$ has a start state $x_s^k$ and a non-empty set of termination conditions $e^k$.

**Termination Criteria –** When an ant is in a state $x$, it checks if one of the termination criteria in $e^k$ is satisfied. If not, it moves to a node $j \in \mathcal{N}^k(x)$. $\mathcal{N}^k$ is a function that returns the neighborhood of a node $x$, i.e. all the nodes on the construction graph $G$ that can be reached from the current state, given the constraints $\Omega$.

**Decision Rule –** An ant chooses the node $j$ according to a probabilistic decision rule, which is a function of the pheromone $\tau$ and the heuristic $\eta$. The specification of these rules is problem dependent, but is usually a random choice biased towards moves with a higher heuristic and pheromone value.

**Pheromone Update –** The pheromone of a path can be modified by an ant as it is traversing it, or on the return journey, when it returns to the start. Again, this is problem dependent, but a standard formulation is to increase the pheromone on good solutions and decrease the pheromone on bad solutions, good and bad being given by the specific formulation.

In terms of algorithmic actions, an ACO algorithm can normally be broken down into three parts. These are:

**ConstructAntsSolutions** This part of the algorithm is concerned with sending ants around the construction graph according to the rules given above.

**UpdatePheromones** This part is concerned with changing the values of the pheromones, by both depositing and evaporating. Parts of this task might be performed during an ant's traversal of the graph, when an ant's traversal is finished or after an iteration of all the ants' traversals.

**DaemonActions** This part of the algorithm performs tasks not directly related to the ants. E.g. a local search procedure might be performed after each ant finishes its traversal.

Given the above framework, multiple artificial ants are released to perform a random walk. This procedure is repeated a number of times, with the pheromone gradually increasing on the best parts of the solution.

There have been many implementations of the above metaheuristic. The first was the original ACO system designed by Dorigo, Maniezzo, and Colorni (1996) known as Ant System. This was used to study the traveling salesman problem, with the construction graph defined by the distances between cities. Another extension to Ant System is the Ant Colony System (ACS) (Dorigo & Gambardella, 1997). Here, the search is biased towards the best-so-far path, with a pseudo-random proportional decision rule that takes the best solution component most of the time and the normal random proportional decision rule the rest of the time. Also, only the best-so-far ant deposits pheromone. ACS is based on a system known as ANT-Q designed by Gambardella and Dorigo (1995), that is itself inspired by the reinforcement learning technique of Q-learning (Sutton & Barto, 1998). The ACS is particularly interesting in this context, as it is the system on which the new work described in later sections has been modeled. This is because this work is inspired by a previous approach to learning Bayesian networks using ACO (described in Section 5.1) which used ACS as its form of ACO.

## 5. Using Ant Colony Optimization in Learning an Equivalence Class

To date, many state-based search algorithms that create a Bayesian network structure have relied on simple techniques such as greedy-based searches. These can produce good results, but have the ever prevalent problem of getting caught at local minima. More sophisticated heuristics have been applied, such as iterated hill climbing and simulated annealing (Chickering, Geiger et al., 1996), but so far, none of these have been applied to E-space. A related approach, by Acid and de Campos (2003) applied tabu search to a space of restricted partially directed acyclic graphs (RPDAGs), a halfway house between the spaces given by DAGs and CPDAGs.

This paper seeks to apply the ACO metaheuristic to E-space, the space of equivalence classes of DAGs. To this end, two extensions are made to the basic metaheuristic. The first is to allow multiple *types* of moves. This is to allow more than one operator to be used in traversing the state space. This is needed, because in general, more than one type of operator is used whilst searching in E-space. The second is to allow the pheromone to be accessed by arbitrary values – normally it is accessed by a single index or two indices. Again this is needed because of the operators used in E-space – the MakeV operator takes three nodes as arguments.

The proposed algorithm, ACO-E, is based in large part on the work of de Campos, Fernández-Luna et al. (2002), which is described in the next section.

## 5.1 Other ACO Algorithms for Learning Bayesian Network Structures

Whilst ACO has been applied to many problems in the area of combinatorial optimization, to date there has not been much research on using the technique to learn Bayesian network structures. Two alternate methods have been defined by de Campos, Fernández-Luna et al. (2002) and de Campos, Gámez, and Puerta (2002). The first conducts a search in the space of orderings of DAGs, whilst the second searches in the space of DAGs. Since a main topic of this work is on this problem, a description of both of these will be given here, in order to examine the early work done on the subject and see how it can inform future studies.

### 5.1.1 ACO-K2SN

In the first technique, known as ACO-K2SN, searching over the space of orderings of DAGs, the various problem components, as taken from Section 4.2 can be defined as follows:

**Construction Graph**  There is one node for each attribute in the data, with an extra dummy node from which the search starts.

**Constraints**  The only constraints are that the tour is a Hamiltonian path.

**Pheromone Trails**  The pheromone is associated with each arc on the graph. Each arc in the graph is intialised to a initial small value.

**Heuristic Information**  The heuristic on each arc is set to the inverse of the negative log likelihood score that is explained below.

**Solution Construction**  The ants work on a system very similar to the ACS system. Beginning at the dummy node, the ants construct a complete path that defines an ordering of the nodes.

**Pheromone Update**  This works exactly as in ACS, with local pheromone updates and global update on the best-so-far solution.

**Local Search**  A version of local search on orderings known as HCSN (de Campos & Puerta, 2001a). This is used on the last iteration of the run.

Given the above components, the search for an ordering proceeds as follows. Starting at the dummy node an ant decides which node to go to next. This will be the first node in the ordering. To choose a node, heuristic information and pheromone is used. The heuristic for the arc from $i$ to $j$ is given by

$$\eta_{ij} = \frac{1}{|f(x_j, Pa(x_j))|},$$

where $f$ is the scoring metric being used and $Pa(x_j)$, the parents of $x_j$ are found by the K2 algorithm, with possible parents being the nodes already visited. The initial pheromone value $\tau_0$ is given by

$$\tau_0 = \frac{1}{n|f(S_{K2SN})|},$$

where $S_{K2SN}$ is the structure given by the K2SN algorithm of de Campos and Puerta (2001b). The update value for the pheromone is given by

$$\Delta\tau_{ij} = \frac{1}{|f(S^+)|},$$

where $S^+$ is the best-so-far structure.

### 5.1.2 ACO-B

The second algorithm given by de Campos, Fernández-Luna et al. (2002) is the ACO-B algorithm. The components for this algorithm are:

**Construction Graph**  There is one node for each possible directed arc between each pair of attributes (excluding self directed arcs). There is also a dummy node that the ants start from.

**Constraints**  The only constraints are that the DAG must be acyclic at each step.

**Pheromone Trails**  The pheromone is associated with each node on the graph. The pheromone at node $(i, j)$ corresponds to the directed arc $j \rightarrow i$.

**Heuristic Information**  The heuristic on each node $(i, j)$ is the gain in score that would occur in adding an arc $j \rightarrow i$.

**Solution Construction**  The ants work on a system very similar to the ACS system. Beginning at the dummy node, the ants construct a path that defines which arcs are added to the DAG. This process ends when there is no gain in score.

**Pheromone Update**  This works exactly as ACS, with local pheromone updates and global update on the best so far solution.

**Local Search**  A standard greedy search with arc addition, deletion and reversal is carried out on the current candidate DAG. This is done every 10 iterations.

As opposed to the ACO-K2SN algorithm given in Section 5.1.1, the search is over the space of DAGs, not orderings of DAGs. Otherwise, there are some similarities in the definitions of parts of the algorithm. The heuristic is given by

$$\eta_{ij} = f(x_i, Pa(x_i) \cup \{x_j\}) - f(x_i, Pa(x_i)),$$

that is, the change in score by adding an arc from $j$ to $i$ in the candidate DAG. The initial pheromone is given by

$$\tau_0 = \frac{1}{n \left| f\left(S_{K2SN}\right) \right|},$$

i.e. it is the same as the heuristic in ACO-K2SN. Also, the pheromone update value is the same as in ACO-K2SN, i.e.

$$\Delta \tau_{ij} = \frac{1}{\left| f\left(S^+\right) \right|}$$

### 5.1.3 PERFORMANCE COMPARISON

In the results given by both de Campos, Gámez et al. (2002) and de Campos, Fernández-Luna et al. (2002), the ACO-B algorithm performs slightly better in terms of accuracy than ACO-K2SN across the ALARM (Beinlich, Suermondt, Chavez, & Cooper, 1989) and INSURANCE (van der Putten & van Someren, 2004) gold-standard networks. It also contains an order of magnitude less statistical tests and so should always be faster. There are more comparisons of ACO-B against other algorithms by de Campos, Fernández-Luna et al. (2002). Here, it is compared against ILS, an iterative local search algorithm with random perturbations of a local maximum and two estimation of distribution (EDA) genetic algorithms, the univariate marginal distribution algorithm (UMDA) by Mühlenbein (1997) and the population-based incremental learning algorithm (PBIL) by Baluja (1994). Compared across the ALARM, INSURANCE and BOBLO (Rasmussen, 1995) networks, ACO-B performed better than the other methods.

## 5.2 Relation of ACO-E to the ACO Metaheuristic

The proposed algorithm, ACO-E, is based in large part, on the work of de Campos, Fernández-Luna et al. (2002). In that work, an ACO algorithm called ACO-B was applied to learning Bayesian networks. This current work differs in that it searches in E-space, uses more than one operator (add an arc) and does not constrain itself to using matrices to store pheromone. The algorithm is shown in Algorithm 1.

In this section, the relation of the various parts of the algorithm to the ACO framework will be given. The problem of learning a Bayesian network structure can be stated as the triple $(\mathcal{S}, f, \Omega)$, where

- $\mathcal{S}$, the set of all candidate solutions, is the set of all CPDAGs on the nodes of the Bayesian network. This set has a massive cardinality, being super-exponential in the number of nodes.

- $f$, the objective function is the function used to score a candidate DAG. This function would generally be one of the scoring criteria mentioned in Section 2.

- $\Omega$, the set of constraints, makes sure that only PDAGs that have consistent extensions are generated as solutions. An explanation of the idea of a consistent extension of a PDAG is given in Section 3.1. In the formulation being presented, the constraints are implicit in the operators that will be used to move from state to state.

Given this statement of the problem, the ACO-E algorithm can be described by the following properties. These properties relate to the ACO metaheuristic described in Section 4.2.

---

**Algorithm 1** ACO-E

---

**Input:** Operators $O$, $t_{max}$, $t_{step}$, $m$, $\rho$, $q_0$, $\beta$, $n$

**Output:** PDAG $\mathcal{P}^+$

   $(\mathcal{P}^+, Path^+) \leftarrow$ GREEDY-E$(\mathcal{P}^{empty}, Path^{empty})$

   $\tau_0 \leftarrow 1/n\,|\text{SCORE}(\mathcal{P}^+)|$

   **for** each operator $o$ in $O$ **do**

      **for** each possible move $m$ in $o$ on $\mathcal{P}^{empty}$ **do**

         $\tau_m \leftarrow \tau_0$

      **end for**

   **end for**

   **for** $t \leftarrow 1$ to $t_{max}$ **do**

      **for** $k \leftarrow 1$ to $m$ **do**

         $(\mathcal{P}^k, Path^k) \leftarrow$ ANT-E$(O, q_0, \rho, \beta, \tau_0)$

         **if** $(t \bmod t_{step} = 0)$ **then**

            $(\mathcal{P}^k, Path^k) \leftarrow$ GREEDY-E$(\mathcal{P}^k, Path^k)$

         **end if**

      **end for**

      $b \leftarrow \arg\max_{k=1}^m \text{SCORE}(\mathcal{P}^k)$

      **if** SCORE$(\mathcal{P}^b) >$ SCORE$(\mathcal{P}^+)$ **then**

         $\mathcal{P}^+ \leftarrow \mathcal{P}^b$

         $Path^+ \leftarrow Path^b$

      **end if**

      **for** each move $m$ in $Path^+$ **do**

         $\tau_m \leftarrow (1-\rho)\,\tau_m + \rho/|\text{SCORE}(\mathcal{P}^+)|$

      **end for**

   **end for**

   return $\mathcal{P}^+$

---

### 5.2.1 The Construction Graph

The construction graph in an ACO algorithm describes the mechanism by which solutions can be assembled. It is specified as the complete graph given over the solution components. As such, these components play a crucial part in the viability of the algorithm.

In the ACO-E algorithm, the components $C$ of the construction graph are the various moves that may be made, i.e. each *move* is an *instantiation* of a supplied operator; in the experiments presented in this paper, the six operators in Table 2 are used. These operators are used as they have been verified to work correctly and effectively by Chickering (2002a). Designing correct operators is difficult, as Chickering showed by finding counter examples to the validity of the operators of Munteanu and Cau (2000). Each ant constructs a solution by walking the construction graph. This corresponds to applying a sequence of moves to a CPDAG. In order for the procedure to begin, a starting state must be specified. In ACO-E this is given as the empty graph.

As usual, the states of the problem are sequences of moves. However, because every state can be a candidate solution, $\mathcal{S} = \mathcal{X}$ in the ACO metaheuristic framework. This does not imply that all states are *feasible* candidate solutions, but only that candidate solutions can be of any length. This also means that $\tilde{\mathcal{S}} = \tilde{\mathcal{X}}$. Another way to view the state of an ant is to consider the empty graph $\mathcal{P}$ (the starting state) and the current state as a sequence of moves (components) $x = \langle c_i, \ldots, c_j \rangle$. Applying each move $c \in x$ in order to $\mathcal{P}$ will give a CPDAG that is another representation of the current state.

It should be noted that the constraints $\Omega$ are implicitly taken care of by the operators, i.e. the validity tests on the operators satisfy the constraint that each state is a valid PDAG. It should also be stated that the usual definition of

$$g(s,t) \equiv f(s,t), \ \forall s \in \bar{\mathcal{S}}$$

applies, and that there is no function $J(x,t)$, since all $x$ are candidate solutions and adding a solution component can decrease the cost.

### 5.2.2 THE PROBLEM HEURISTIC

In an ACO algorithm, the heuristic is used to guide the search to good solutions. It often does this implicitly in terms of a cost associated with choosing a particular component to add to the current state; adding a component with the least cost is often a useful way of proceeding in constructing a solution.

In ACO-E, the heuristic is used in the same manner, with the addition that the cost for adding a component can be negative, i.e. adding a component to the current state can improve the cost function $g$. The heuristic is dynamic in that it depends on the current state of the ant. Also, it is associated with each component $c \in C$ as opposed to the arcs $c_i - c_j$ between components.

The value of the heuristic $\eta_i$ is given by the score gain for each move $c_i \in C$ that is possible given the current state. In essence it corresponds to the change in score given by performing a particular move on the current CPDAG. For the operators being used in this article, this means the values in Table 2.

### 5.2.3 THE PROBLEM PHEROMONE

The pheromone in an ACO algorithm guides the search based on the results of previous searches. In many instances, it is associated with the arcs on the construction graph, but in ACO-E it is associated with the nodes of the construction graph. This gives pheromone values $\tau_i$ for each $c_i \in C$.

The pheromone for each $\tau_i$ is initialised to a value $\tau_0$ given by

$$\tau_0 = \frac{1}{n \left| \text{SCORE}\left( \mathcal{P}^+ \right) \right|}. \tag{2}$$

In this formula, $n$ is the number of variables that are in the data, SCORE is the objective function $f$, as defined in Section 5.2 and $\mathcal{P}^+$ is the best-so-far solution. At the start of the algorithm, this is initialised to that found by a greedy search starting from the empty graph.

In order that the pheromone may change to reflect the tours of ants, pheromone update rules are given. Similar to ACS, there is a local evaporation rule, whereby pheromone is removed from a path as an ant traverses it

$$\tau_m \leftarrow (1 - \rho) \tau_m + \rho \tau_0$$

This shows the effect of the parameter $\rho$, which is the pheromone evaporation and deposition rate. With this formula, there are implicit bounds on how high and low the pheromone at each component can get. Also similar to ACS, there is a global pheromone update rule that deposits new pheromone on the best-so-far path

$$\tau_m \leftarrow (1 - \rho) \tau_m + \rho / \left| \text{SCORE}\left( \mathcal{P}^+ \right) \right|$$

This occurs at the end of a run of ants. Again, SCORE and $\mathcal{P}^+$ are defined as in Equation 2. Also again, this formula implements implicit limits on the values that pheromone can take.

### 5.2.4 PROBABILISTIC TRANSITION RULE

In choosing which component to visit next given a particular state, an ACO algorithm utilises a probabilistic transition rule. This rule normally uses values given by the heuristic and pheromone to inform the choice of which node to pick. The actual choice is random and is based on a distribution given by the heuristic and pheromone of each possible choice. In ACO-E, the probabilistic choice rule is given by a pseudo random proportional choice rule, very similar to the one used in ACS. This type of rule allows the balance between exploration and exploitation to be varied. Being able to change this balance is important, as it has been shown to produce quite different results (Dorigo & Stützle, 2004). An ant chooses component $c_m$, where $m$ is given by

$$m \leftarrow \begin{cases} \arg\max_{m \in \mathcal{N}(x)} \tau_m \left[\eta_m\right]^\beta, & \text{if } q \leq q_0 \\ \text{random proportional,} & \text{otherwise.} \end{cases}$$

In this formula, $\mathcal{N}(x)$ is the set of components that an ant at state $x$ can move to, given the problem constraints $\Omega$. The rule is *pseudo*-random proportional, because it sometimes behaves in a manner that is not random. A random number $q$ is drawn uniformly in the range $[0, 1]$. If this number is less than or equal to a parameter $q_0$, then the rule behaves greedily; the best move possible is taken dependent on the value of $\tau_m \left[\eta_m\right]^\beta$ for each component $c_m$. Here, $\tau_m$ and $\eta_m$ are the pheromone and heuristic as explained previously and $\beta$ is a parameter that says how much to favour the heuristic over the pheromone.

If the number $q$ is greater than $q_0$ than a random proportional rule is used to select which component to visit next. The probability that the ant will visit component $c_m$ is given by $p_m$, where

$$p_m = \frac{\tau_m \left[\eta_m\right]^\beta}{\sum_{\mu \in \mathcal{N}(x)} \tau_\mu \left[\eta_\mu\right]^\beta}, \quad \forall m \in \mathcal{N}(x). \tag{3}$$

It can be seen that the probability that an ant moves to component $c_m$ is directly given by $\tau_m \left[\eta_m\right]^\beta$, normalised over the other possible moves so that it is in the range $[0, 1]$.

### 5.2.5 PROPERTIES OF ANTS

In terms of the ants used to construct solutions, the following properties of ant $k$ should be noted:

- The memory $\mathcal{M}^k$ can be equated to the current state of the problem given by ant $k$. From this, the current CPDAG can be constructed in order to implement the constraints $\Omega$, compute the heuristic values $\eta$, evaluate the current solution and lay pheromone on the tour. In practice, the current CPDAG is normally kept in order to avoid having to recompute it at every step.

- The start state $x_s^k$ is given by the empty sequence $\langle \rangle$, i.e. the empty CPDAG.

- The single termination condition $e^k$, is to stop the tour when no improvement in score is possible.

- The neighborhood $\mathcal{N}^k(x)$ is the set of all valid moves given the current CPDAG.

### 5.2.6 LOCAL SEARCH PROCEDURE

As is often the case with ACO algorithms, ACO-E can use a local search procedure at intermediate points throughout the run of the algorithm and at the end. This local search procedure can be used to quickly bring a solution to a local maximum. With the current heuristic and the standard local search that would be used in these circumstances – greedy search with the operators defined in Table 2, known here as GREEDY-E – local search would provide no additional benefit over the solution found by an ant. Nevertheless, the local search was put in the algorithm in the case that the problem heuristic was implemented differently. An example of this would be a static heuristic obtained by scoring operations on an empty graph. Since this is invariant over the algorithm run, it would only need to be calculated once at the start of the run.

## 5.3 Description of ACO-E

This section will focus on giving an algorithmic description of ACO-E. This is done in conjunction with the pseudo code given in Algorithms 1 and 2. ACO-E takes as input a number of parameters and returns the best PDAG found, according to a scoring criterion SCORE, that is defined as the objective function $f$. It is assumed that scoring criteria generally give negative values; the higher the value, the better the model. This is the case of most of the standard criteria as discussed in Section 2. The meaning of the parameters is as follows:

$O$  This is a set of operators that can modify the current PDAG state in the search. Examples of these are the ones given in Table 2, e.g. InsertU, DeleteU, etc. However, other operators could be used, e.g. those of Munteanu and Cau (2000) and Munteanu and Bendou (2001).

$t_{max}$  This is the number of iterations of the algorithm to run. At each iteration, a number of ants construct solutions. Pheromone deposition happens after all the ants have finished their tours.

$t_{step}$  This is the gap, in iterations, between which local search procedures are run. If set so that $t_{step} > t_{max}$, then local search only happens at the end of the algorithm run.

$m$  This is the number of ants that run at each iteration.

$\rho$  This, a value in $[0, 1]$, is the rate at which pheromone evaporates and is deposited. It is used in both the pheromone evaporation and pheromone deposition rules in Section 5.2.3.

$q_0$  This, a value in $[0, 1]$, gives the preference of exploitation over exploration. It is used in the pseudo-random probabilistic transition rule as explained in Section 5.2.4.

$\beta$  This exponent gives the relative importance of the heuristic over the pheromone levels in deciding the chance that a particular trail will be followed. It is used in the pseudo-random probabilistic transition rule in Section 5.2.4.

$n$  This is the number of nodes in the PDAG.

There are also other variables in the algorithm. These include:

$\mathcal{P}^+$  the best-so-far PDAG;

$Path^+$  the best-so-far path;

---

**Algorithm 2** ANT-E

---

**Input:** Operators $O$, $\rho$, $q_0$, $\beta$

**Output:** PDAG $\mathcal{P}$, Path *Path*

 Empty PDAG $\mathcal{P}$, Empty path *Path*

 **while** true **do**

  $M \leftarrow$ All possible moves from $\mathcal{P}$ using $O$

  **if** $|M| = 0 \lor \max_{l \in M}$ TOTAL-SCORE$(l, \beta) \leq 0$ **then**

   return $(\mathcal{P}, Path)$

  **end if**

  $q \leftarrow$ random number in $[0, 1)$

  **if** $q \leq q_0$ **then**

   $l \leftarrow \arg\max_{l \in M}$ TOTAL-SCORE$(l)$

  **else**

   $l \leftarrow$ random according to Equation 3

  **end if**

  $\tau_m \leftarrow (1 - \rho)\,\tau_l + \rho\,\tau_0$

  $\mathcal{P} \leftarrow$ apply $l$ to $\mathcal{P}$

  *Path* $\leftarrow$ append $l$ to *Path*

 **end while**

---

$\mathcal{P}^{empty}$ the empty PDAG; and

$Path^{empty}$ the empty path, i.e. the path with no entries.

In starting the algorithm, a greedy search (called GREEDY-E) is performed. This is a search through the space of equivalence classes using the framework and operators given by Chickering (2002a) and shown in Table 2. It gives a starting best-so-far graph and path from which the search can proceed. Pheromone levels for each solution component are then initialised to $\tau_0 = 1/n\,|\text{SCORE}\,(\mathcal{P}^+)|$. The main loop of the algorithm then begins for $t_{max}$ iterations. At each iteration, $m$ ants perform a search, given by algorithm ANT-E, shown in Algorithm 2. Also, for every $t_{step}$ iterations, a local search is performed on the PDAGs returned from ANT-E, to try and improve results. Using local search as part of an ACO algorithm is a very common technique (Dorigo & Stützle, 2004), as it is a easy way to obtain good results with little effort. After the $m$ ants have traversed the graph, the best graph

---

**Algorithm 3** TOTAL-SCORE

---

**Input:** Move $l$, $\beta$

**Output:** Score $s$

**return** $s$ such that $s = \begin{cases} \tau_l\,(\eta_l)^\beta & \text{if } \eta_l > 0 \\ 0 & \text{otherwise} \end{cases}$

---

and path are selected from the best-so-far graph and path and the ones found by each of the ants in the current iteration. Finally, the global pheromone update lays and evaporates pheromone on the best-so-far path.

 The ANT-E algorithm creates a PDAG by examining the various states that may be proceeded to from the current state, given a set of operators that may act on the current PDAG. It then selects a new state based on a random-proportional choice rule. The parameters to the function have the same description as the ones to the ACO-E function.

Figure 7: Bayesian network used in sample trace

Starting out, the algorithm constructs an empty PDAG. Then at each stage a move is made to a new PDAG, which can be reached by applying one of the operators in $O$. Initially, a number is given to each move by TOTAL-SCORE, shown in Algorithm 3. This number represents a weight given to each move $l$ depending on the current pheromone associated with making that move $\tau_l$, and the heuristic associated with making the move $\eta_l$. This heuristic is given by the increase in score obtained by taking that move, higher overall scores meaning better solutions. If there can be no increase in the score, the ant stops and returns the solution $\mathcal{P}$ and the path followed. Otherwise there is a possible move and the ant decides how to make it. Firstly a random number $q$ is obtained. If it is less than a specified value $q_0$, then the best move is taken. If it is greater than $q_0$, then a random proportional choice is made, with the probability of better moves being higher. After this, a local pheromone update is applied to the path just taken, the path is updated with the new location at the end and the current state is updated to become the new state given by $l$. Note that applying a move to a CPDAG to change state implies that the resulting PDAG will be extended to a DAG by a suitable method (e.g., that of Dor & Tarsi, 1992) and this DAG be changed back to a CPDAG. Details can be found in the article of Chickering (2002a).

## 5.4 Trace of Algorithm Execution

As a simple example of the execution of the ACO-E algorithm, a trace of its behavior during an actual execution will be given during this section. Consider the Bayesian network in Figure 7. This network is fully specified, with a DAG structure and parameters given in the form of conditional probability tables. As can be seen, the variable 0 can take on the values a and b, the variable 1 can take on the values c and d and the variable 2 can take on the values e, f and g.

| $\rho$ | $q_0$ | $\beta$ | $t_{max}$ | $m$ |
|--------|-------|---------|-----------|-----|
| 0.1    | 0.1   | 1.0     | 1         | 2   |

Table 3: Parameters for sample trace

|   | Move | $\tau$ | $\eta$ |
|---|------|--------|--------|
|   | InsertU(0,1) | 0.00312 | -0.21565 |
| 1 | InsertU(0,2) | 0.00312 | 34.6527 |
|   | InsertU(1,2) | 0.00312 | 11.2204 |
|   | InsertU(0,1) | 0.00312 | -0.21565 |
| 2 | InsertU(1,2) | 0.00312 | 11.2204 |
|   | DeleteU(0,2) | 0.00312 | -34.6527 |
|   | InsertU(0,1) | 0.00312 | 0.37742 |
| 3 | DeleteU(0,2) | 0.00312 | -34.6527 |
|   | DeleteU(1,2) | 0.00312 | -11.2204 |
|   | MakeV(0,1,2) | 0.00312 | 0.59307 |

|          | Move | $\tau$ | $\eta$ |
|----------|------|--------|--------|
|          | InsertU(0,1) | 0.00312 | -0.21565 |
|          | DeleteD(0,2) | 0.00312 | -35.2457 |
| 4 (Ant 1)| DeleteD(1,2) | 0.00312 | -11.8134 |
|          | ReverseD(0,2) | 0.00312 | -0.59306 |
|          | ReverseD(1,2) | 0.00312 | -0.59307 |
|          | DeleteU(0,1) | 0.00312 | -0.47742 |
| 4 (Ant 2)| DeleteU(0,2) | 0.00312 | -35.2457 |
|          | DeleteU(1,2) | 0.00312 | -11.8134 |

Table 4: Values corresponding to the moves in Figure 8

For the purposes of this demonstration, 90 data were sampled from this Bayesian network. The ACO-E algorithm was then started with the parameters set as in Table 3. The PDAG found from the initial GREEDY-E run was the same as the sample Bayesian network structure. $\mathcal{P}^+$ was then set to this PDAG. The score of this PDAG was 106.918. $\tau_0$ was then set to 0.00312. Because $t_{max}$ was set to 1, there was only one iteration of the algorithm. On this iteration, two ants constructed solutions using the ANT-E procedure. The trace of how these ants proceeded is shown in Figure 8 and Table 4. In the diagram, the sequence of moves can be seen along with the value of $q$ at each step. The score of the final network for each ant is also shown. In the table, the possible moves at each point for each ant are shown, along with the pheromone $\tau$ and heuristic value $\eta$. It should be noted that the pheromone is the same for all moves, as this was the very start of the ACO-E algorithm and no pheromone deposition had occurred. At each move, pheromone evaporation occurs, but once more, no difference is found because all the pheromone values are equal to $\tau_0$.

After the two ants finish their run, the best solution is chosen as variable $b$. In this case it is that of Ant 1, with a score of -106.918. This is then compared to the score of $\mathcal{P}^+$. Because the two structures are the same, there is no score difference and hence no change occurs. Pheromone deposition then occurs on the moves that made up $\mathcal{P}^+$, i.e. the moves in $Path^+$. In this case, the pheromone for InsertU(0,2), InsertU(1,2) and MakeV(0,1,2) got updated to $(1-0.1) \cdot 0.00312 + 0.1/|-106.918| = 0.00374$. Since $t_{max}$ was set to 1, there are no more iterations and the algorithm returns $\mathcal{P}^+$.

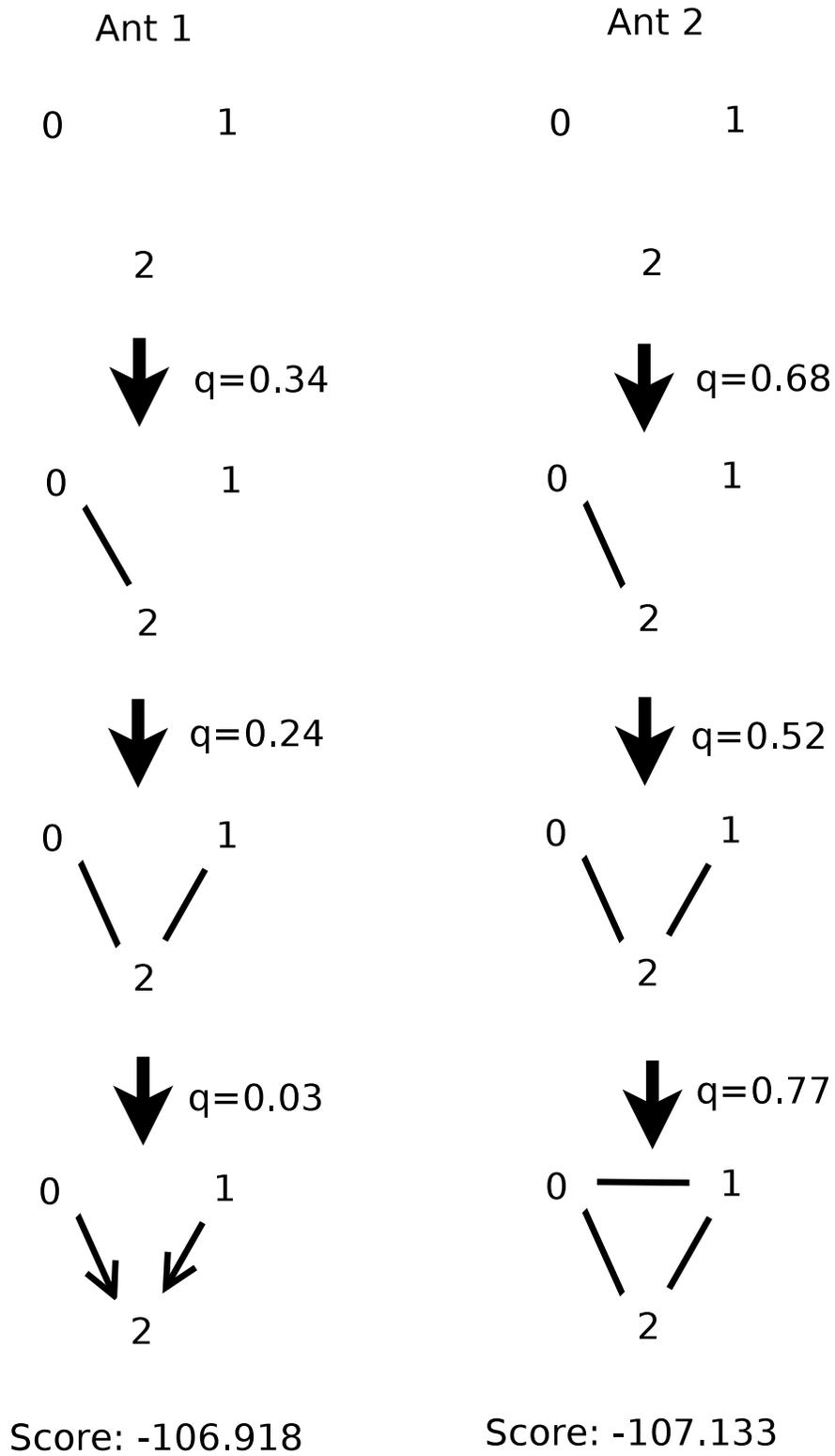

Figure 8: Trace of progress in ANT-E

### 5.5 Implementation Issues

In implementing the algorithms given in this paper, care must be taken to avoid long run times. Firstly, caching the score of a node given it's parents is a simple technique that can greatly improve performance. Secondly, caching the results of the validity tests needed to check which moves are applicable at a certain state, can again increase performance dramatically. However this technique is not as easy to implement as it might appear (Daly, Shen, & Aitken, 2006).

Care must also be taken in implementing the pheromone for the moves. Traditionally, matrices of values are used, which allow fast access and updating. However in the case of the MakeV operator, which takes three indices, a three dimensional matrix would be needed. This would quickly become infeasible as the problem size grew, especially as only some of the entries would be used. This would be due to the algorithm never getting to those states. Instead a structure such as a map can store this information. A map can scale linearly with the number of elements actually being used. If the map is implemented as a tree, entries can be accessed in logarithmic time and if a hash table is used, access can be in constant time.

## 6. Experimental Methodology

This section is concerned with testing the ACO-E algorithm presented in Section 5 and the evaluation of the results produced. In order to facilitate understanding of the experimental methodology used, the section will be structured as follows.

Firstly, an account will be given of the objects on which the testing will be performed. These objects are six gold-standard Bayesian networks that are well known in the field. The various properties of the networks will be discussed. From these networks data can be sampled and it is this data that can be used as input to the algorithms.

Then, experiments using the ACO-E algorithm will be shown. The methodology used in running the experiments will be defined, along with a description of the various evaluation criteria. These involve criteria well known in the field. Two different sets of experiments will be presented, one focused on the comparison of ACO-E against similar algorithms, the other a comparison of ACO-E against state-of-the-art algorithms. Also, the behavior of the ACO-E algorithm for different parameters will be shown.

### 6.1 Standard Bayesian Networks

In this section a set of six gold-standard Bayesian networks will be presented. These networks will be the basis of the testing that will be showcased later. Various properties of the networks will be given, covering: the number of nodes of the structure, the number of edges in the structure, the average number of in edges etc.

#### 6.1.1 Six Gold-Standard Networks

In the experiments shown in the next section, six gold-standard networks are used. These are the ALARM (Beinlich, Suermondt et al., 1989), Barley (Kristensen & Rasmussen, 2002), Diabetes (Andreassen, Hovorka, Benn, Olesen, & Carson, 1991), HailFinder (Abramson, Brown, Edwards, Murphy, & Winkler, 1996), Mildew (Jensen, 1995) and Win95pts networks (Microsoft Research, 1995). These networks were chosen because they covered a wide range of domains, were easily available and all contained discrete attributes. The last property was important because the scoring criterion that would be used in the experiments is implemented over multinomial random variables.

|              | Alarm | Barley | Diabetes | HailFinder | Mildew | Win95pts |
|--------------|-------|--------|----------|------------|--------|----------|
| Nodes        | 37    | 48     | 36       | 56         | 35     | 76       |
| Edges        | 46    | 84     | 48       | 66         | 46     | 112      |
| Mean In-Degree | 1.24 | 1.75  | 1.33     | 1.18       | 1.31   | 1.47     |
| V-Structures | 26    | 66     | 21       | 37         | 37     | 135      |
| V-Struct/Nodes | 0.70 | 1.38  | 0.58     | 0.66       | 1.06   | 1.78     |

Table 5: Bayesian network properties

Various properties of these Bayesian networks are shown in Table 5. In this table, *Nodes* and *Edges* specify the number of nodes and edges respectively in the graph. The *Mean In-Degree* is the average number of arcs coming into a node in the graph. This is equal to the Mean Out-Degree and the number of edges divided by the number of nodes. Finally, *V-Structures* and *V-Struct/Nodes* show the amount of v-structures in the graph and the amount of v-structures divided by the number of nodes.

## 6.2 Methodology

This section contains details of the experiments performed using the ACO-E algorithm described in Section 5. Firstly, the methodology used in running the experiments will be presented. This includes an analysis of the needed outcomes, the design of five experimental conditions and an explanation of the evaluation criteria.

### 6.2.1 EXPERIMENTAL DESIGN

In designing an experimental methodology to test the efficacy of the ACO-E algorithm, three different outcomes were desired.

- The first was to analyze the behavior of the algorithm as a function of the parameters and the test networks. This is needed in order to try and understand the range of values in which parameters might be useful and to show the effect of the ACO behavior on outcomes.

- The next desired outcome was to test ACO-E against other similar algorithms. To this end, ACO-E was tested against another ACO algorithm and algorithms that searched in the space of equivalence classes.

- Finally the last desired outcome was to test ACO-E against state-of-the-art algorithms from the literature. These tests would show the comparative usefulness of ACO-E against other well-known and good-performing methods.

In order to obtain these outcomes, various experimental conditions were designed, which will be explained below.

**The Scoring Function**    For these experiments, it was decided to use the BDeu criterion invented by Buntine (1991) and described in Section 2. According to the study by Shaughnessy and Livingston (2005), BDeu had the best tradeoff between precision and recall of edges (confusingly BDeu is called BAYES in their study, with BDeu in their study meaning the K2 metric). This criterion gives a fully Bayesian score, with the assumption of Dirichlet parameter priors and a uniform prior over all possible states of the joint distribution given the prior network. To fully specify the BDeu criterion, two pieces of information are needed. First is a prior on structures $P(\mathcal{G})$. This could be a uniform prior, such that all structures have the same $P(\mathcal{G})$. Another method shown by Heckerman, Geiger et al. (1995) was to have an expert specify a structure, and have a method that penalises differences between the expert's structure and a candidate structure.

The second piece of information needed is the 'equivalent sample size', $N'$, a parameter that encodes the confidence in the prior parameters and prior structure. Selecting this value can be troublesome (Silander, Kontkanen, & Myllymaki, 2007; Steck & Jaakkola, 2003), but 'reasonable' values in the range $[1, 10]$ often work well.

In recognition that simpler structures are often more appealing, the prior was specified by the method shown by Heckerman, Geiger et al. (1995). In their formulation, two objects are specified; a prior structure $\mathcal{G}_{prior}$ and the prior distribution given as:

$$P(\mathcal{G}) = c\kappa^{\delta},$$

where $c$ is a normalisation constant that can be ignored, $\kappa$ is a parameter that needs to be specified and $\delta$ is given by the formula

$$\delta = \sum_{i=1}^{n} \delta_i,$$

where $\delta_i$ is the symmetric difference of the parent set for node $i$ between $\mathcal{G}_{prior}$ and $\mathcal{G}$.

**Condition 1** Experimental condition 1 was designed to analyze the behavior of ACO-E across different parameters and to compare against other similar algorithms. These algorithms were ACO-B (de Campos, Fernández-Luna et al., 2002), EPQ (Cotta & Muruzábal, 2004; Muruzábal & Cotta, 2004) and a greedy search in the space of equivalence classes using Chickering's operators (Chickering, 2002a) (called GREEDY-E here). A description of these will now be given.

**ACO-B** ACO-E is based in part on the construction of this algorithm and so there are some similarities. ACO-B is an ACO based algorithm that provides a search through the space of DAGs, with each of its moves being the addition of a directed arc to the current DAG. A more detailed description is given in Section 5.1.2.

**EPQ** This method uses an evolutionary programming algorithm that performs a search over the space of equivalence classes of DAGs. Like Chickering (2002a), they explicitly use CPDAGs (defined in Section 3.1) to represent the individuals, i.e. equivalence classes of DAGs. At each generation, from a population of $P$, members of the population are selected using a binary tournament and mutated using the operators of Chickering. The best $P$ out of the $2P$ selected are then put forward into the next round, for $T$ rounds.

**GREEDY-E** This algorithm uses the operators of Chickering to perform a greedy search in the space of CPDAGs. The results of tests performed by Chickering showed that the search generally performed better than search in the space of DAGs.

| Parameter | Value |
|-----------|-------|
| $N'$ | 4 |
| $\kappa$ | 0.2 |
| $t_{max}$ | 200 |
| $m$ | 5, 7, 10, 12, 15, 20 |
| $\rho$ | 0.0, 0.1, 0.2, 0.3, 0.4, 0.5 |
| $q_0$ | 0.7, 0.75, 0.8, 0.85, 0.9, 0.95 |
| $\beta$ | 0.0, 0.5, 1.0, 1.5, 2.0, 2.5 |

Table 6: Parameter values for testing ACO-E

For the experiments in this section, testing involved the six standard networks presented in Section 6.1.1. The BDeu scoring criterion was used, and as suggested by Kayaalp and Cooper (2002) and by Heckerman, Geiger et al. (1995), an equivalent sample size of 4 was used for the parameter priors. Also an empty structure prior with $\kappa$ as defined by Heckerman, Geiger et al. (1995) was used. For each individual run, 10,000 data were sampled from the network and used to construct the scoring function. Then for each combination of values for the parameter settings of $\rho$, $q_0$, $\beta$ and $m$, a run of the experiment was made for both the ACO-E and ACO-B algorithms. The range of values that these parameters were taken from are shown in Table 6.

In total this gave 1296 runs for each algorithm, for each network. As a consequence, this gave a total of 216 results for each setting of a parameter. In order to match this number of runs, the EPQ and GREEDY-E algorithm were also run 216 times each. It should be stressed that each run of ACO-E using a particular combination of parameters and each run of EPQ and GREEDY-E was done with a different data set sampled from the network. This technique guards against overfitting the parameters to a particular data set. It should also be noted that for each algorithm, a limit of 5 parents was allowed for a node, in order to speed up algorithm execution.

**Condition 2**   Experimental condition 2 was designed to test ACO-E against other state-of-the-art Bayesian network structure learning algorithms. For these purposes the results found in the study conducted by Tsamardinos, Brown, and Aliferis (2006) was used. This study produced a thorough comparison of many different algorithms and made the results available, which allows the results for ACO-E to be compared against all of the algorithms used in the study. The various parameters used for ACO-E (that had equivalent parameters in other algorithms) were kept as close as possible to those used by Tsamardinos, Brown et al. The various algorithms that were compared against were: the max-min hill-climbing algorithm (MMHC)  (Tsamardinos, Brown et al., 2006), the optimal reinsertion algorithm (OR)  (Moore & Wong, 2003), the sparse candidate algorithm (SC)  (Friedman, Nachman, & Pe'er, 1999), a greedy search using the three standard operators as in Table 1 (GS), the PC algorithm (PC)  (Spirtes, Glymour et al., 2000), the three phase dependency analysis algorithm (TPDA)  (Cheng, Greiner, Kelly, Bell, & Liu, 2002) and the greedy equivalent search algorithm (GES)  (Chickering, 2002b).

For these experiments, testing involved four of the six standard networks presented in Section 6.1.1; Alarm, Barley, HailFinder and Mildew. These networks were used as the experiments of Tsamardinos, Brown et al. did not use the other two (Diabetes and Win95pts). The other networks shown in the paper of Tsamardinos, Brown et al. were not used as they were not available in a usable format. For each run of the algorithm, 5000 data were generated by sampling the particular networks in question. This was chosen as opposed to the 10,000 data in Condition 1, as this was the amount chosen by Tsamardinos, Brown et al.

| Parameter | $N'$ | $\kappa$ | $t_{max}$ | $m$ | $\rho$ | $q_0$ | $\beta$ |
|-----------|------|----------|-----------|-----|--------|-------|---------|
| Value | 10 | 0.09 | 200 | 20 | 0.4 | 0.75 | 0.75 |

Table 7: Parameter values for testing ACO-E

| | Alarm | Barley | Diabetes | HailFinder | Mildew | Win95pts |
|-----------|-------|--------|----------|------------|--------|----------|
| $\rho$ | 0.4 | 0.4 | 0.4 | 0.2 | 0.4 | 0.2 |
| $q_0$ | 0.8 | 0.8 | 0.7 | 0.8 | 0.7 | 0.95 |
| $\beta$ | 0.5 | 1.0 | 1.0 | 1.0 | 0.5 | 2.5 |

Table 8: Tuned parameters for ACO-E

As in Condition 1, the BDeu scoring function was used. The parameter values of this function and the ACO-E parameters are shown in Table 7. The ACO-E parameter values were chosen as they represented reasonable values that should perform well on most instances. Each experiment was run 100 times for each network.

**Condition 3**  Condition 3 was designed with a number of objectives in mind. These were:

- examine the effect of different sample sizes on ACO-E output;

- use a separate test sample in scoring networks output from ACO-E; and

- examine the complexity of ACO-E by noting the number of statistics computed during a run.

In order to achieve these objectives, new experiments were run. In these experiments, the parameters were set by examining the output of the experiments of Condition 1 – these outputs can be seen in Section 7.1. The optimum value for the parameters was chosen by finding the best combination from Condition 1 (note that Table 10 shows the average BDeu score for each parameter setting). The experiments were performed across the six standard networks, with five different sample sizes – 100, 500, 1000, 5000 and 10000. The various parameters were set as in Table 8.

Each combination of network and sample size was run 100 times. The various other parameters were set as $m = 20$ and $t_{max} = 200$. The BDeu scoring criterion was used, with an empty structure prior, an equivalent sample size $N'$ of 4 and a value of $\kappa = 0.05$. The meaning of the BDeu parameters has been described above.

**Condition 4 – Tuned Metaheuristics**  In order to be able to compare ACO-E to the other metaheuristics described in Condition 1, experiments were run with tuned parameters. The experiments were performed across the six standard networks, with a sample size of 10000. For ACO-B, the various parameters were set as in Table 9. These combination of parameters gave the best BDeu score for ACO-B in Condition 1. GREEDY-E and EPQ have no meaningful parameters to tune.

Each experiment was run 100 times. For ACO-B, the various other parameters were set as $m = 20$ and $t_{max} = 200$. Similar to Condition 3, the BDeu scoring criterion was used, with an empty structure prior, and equivalent sample size $N'$ of 4 and a value of $\kappa = 0.05$. In these runs, a limit of 7 parents was allowed for a node, as opposed to the 5 of Condition 1.

|         | Alarm | Barley | Diabetes | HailFinder | Mildew | Win95pts |
|---------|-------|--------|----------|------------|--------|----------|
| $\rho$  | 0.1   | 0.5    | 0.5      | 0.4        | 0.4    | 0.1      |
| $q_0$   | 0.85  | 0.7    | 0.8      | 0.9        | 0.7    | 0.95     |
| $\beta$ | 2.0   | 2.0    | 2.0      | 2.0        | 2.5    | 2.5      |

Table 9: Tuned parameters for ACO-B

**Condition 5 – Examining the Applicability of ACO-E**   Experimental Condition 5 was designed to test the applicability of ACO-E to given data sets. To achieve this, a simple procedure was designed to indicate to what level the ACO-E algorithm would perform better than a simple greedy search. This procedure is based on the GREEDY-E algorithm mentioned in Condition 1. The procedure is as follows.

An original data set is sampled with replacement and the GREEDY-E algorithm is run. For the purposes of these experiments, this original data set was sampled from a Bayesian network. When the algorithm terminates, the number of v-structures in the returned structure is counted and divided by the number of variables in the data set. This statistic is noted and the procedure starts again, with a new set of resampled data. The whole procedure is repeated until a confident prediction of the normalized v-structure mean can be made. The mean value obtained can be used as a measure of the complexity of the search space. A higher value indicates more v-structures and hence a more complicated space.

For the purposes of this paper, the BDeu scoring function with an equivalent sample size $N'$ of 4 and equal structure priors was used. Test were performed across each of the six standard networks and at sample sizes of 100, 500, 1000, 5000 and 10000. 100 resamplings were used in each case.

### 6.2.2 Evaluation Criteria

In the running of these experiments, various scoring metrics were picked to ascertain how well certain algorithms behaved. These were: the scoring function used in running the experiments, a test scoring function that was based on a different sample, the structural Hamming distance (SHD), the number of scoring function evaluations and the number of distinct scoring function evaluations. These are explained below.

**The Scoring Function**   For all experiments, the BDeu scoring function was used with differing parameters, depending on the experimental condition. Because these parameters were uniform given the condition, the score value of a Bayesian network structure could be used to compare the results of different algorithms. In terms of the BDeu score, this means that the higher the average score achieved, the better the results.

**Test Scoring Function**   As well as the scoring function used in the running of the algorithm, a separate BDeu scoring function was defined, using an independent, same-size sample from the network being used.

**Structural Hamming Distance**    In order to provide an objective measure of network structure reconstruction behavior and to compare results against the work of Tsamardinos, Brown et al. (2006), the value of the structural Hamming distance (SHD) metric is given. This measures the difference between the learned network and the gold-standard generating network. Both networks are transformed from DAG to CPDAG (if not already in this representation) and penalties are given for the number of missing and extra edges and for incorrectly directed arcs.

**Score Function Evaluations**    In order to estimate the complexity of running the ACO-E algorithm, two statistics were measured. The first statistic is the number of times the scoring function has been evaluated up to a particular point in time.

**Distinct Score Function Evaluations**    The next statistic is the number of times a *distinct* scoring function evaluation has occurred, i.e. the number of times the arguments to the scoring function are different. This statistic is often wildly different to the total number of scoring function evaluations and is often a better measure of complexity, as caching of evaluations is a standard technique to speed up algorithm runs.

## 7. Experimental Results

In this section the results of experiments performed according to the methodologies given in Section 6.2 will be presented. In 6.2, five experimental conditions were given. The first dealt with analyzing the behavior of ACO-E with respect to its parameters and in comparison to other metaheuristic algorithms that shared similar behavior. The second condition dealt with comparing ACO-E to other state-of-the-art Bayesian network structure learning algorithms. The third condition focused on the effect of sample sizes on output quality, the behavior of a scoring function defined on a separate test set and the computational complexity of the algorithm. The fourth looked at the behavior of the metaheuristic algorithms with tuned behavior. Finally the fifth condition dealt with the situations when ACO-E should be used. These results will be presented in this order, followed by a discussion and interpretation of these results.

### 7.1 Condition 1

The results of the runs using experimental condition 1 are shown in two sets, which reflect how they will be analyzed later. Firstly, detailed results for ACO-E are shown in Tables 10 and 11. In these tables, the figures given are the results over all other parameters; e.g. the figure for $\rho = 0.1$ is given by calculating the mean and standard deviation over all results with $\rho = 0.1$. In this case, the size of the samples will be 216, and will be calculated over all combinations of the other parameters. It should be noted that the specific values of $\rho = 0$ and $\beta = 0$ are special cases. When $\rho = 0$, there is no pheromone evaporation and no pheromone deposition on the graph; i.e. pheromone plays no part in the algorithm. With $\beta = 0$, there is no heuristic used whilst the ants traverse the construction graph.

The comparative results involving ACO-E, ACO-B, GREEDY-E and EPQ are shown in Table 12 and Figures 9 and 10. These show the behavior of ACO-E against other algorithms, both as a function of the algorithm iteration and as a final value. In these results, the iterations figure is that for ACO-E and ACO-B. The EPQ iteration number is three times that of the shown iteration. As such, whilst ACO-E and ACO-B were run for 200 iterations, EPQ was run for 600 and the results scaled to 200. This can be done, as the concept of an iteration in one framework does not translate well in terms of time to another framework.

## 7.2 Condition 2

The results of the experiments conducted to experimental Condition 2 are illustrated here. The second set of comparisons involved ACO-E against other state-of-the-art Bayesian network structure learning algorithms.

The results of this comparison are shown in Table 13. The acronyms specified are as given by Tsamardinos, Brown et al. (2006) and were discussed before in Section 6.2.1. Some of the results as supplied by Tsamardinos, Brown et al. are missing and are marked by 'N/A' in Table 13. If a result is out of the range of most others, it is represented as a number stating the median.

## 7.3 Condition 3

The results of the experiments conducted according to Condition 3 are shown here. This set of experiments was designed to show the effects of sample size on ACO-E output and also provide a measure of the computational complexity of the algorithm.

The SHD results of the runs after 200 iterations can be seen in Table 14, whilst the score results after 200 iterations are in Table 15. Table 16 shows the score results from a different test sample.

The remaining results from these experiments are shown in Figures 11 and 12. These show the total number of score evaluations and distinct number of score evaluations respectively, for runs of the ACO-E algorithm.

## 7.4 Condition 4

Experimental Condition 4 was used to compare ACO-E against the metaheuristic algorithms used in Condition 1 when the parameters had been tuned to the best combinations from Condition 1. The other algorithms were ACO-B, EPQ and GREEDY-E. These results are consolidated into Table 17 which show the results after the runs have finished.

## 7.5 Condition 5

The results of the experiments under Condition 5 are shown in Table 18. With these experiments, multiple searches were performed using the GREEDY-E algorithm, with data being resampled for each experiment. Experiments were performed 100 times across all combination of the test networks and sample sizes.

| $\rho$ | | | | | |
|---|---|---|---|---|---|
| 0.0 | 0.1 | 0.2 | 0.3 | 0.4 | 0.5 |
| Alarm $(-10^5)$ 1.0383 ± 0.0037 | 1.0385 ± 0.0036 | 1.0387 ± **0.0035** | 1.0385 ± 0.0037 | 1.0388 ± 0.0037 | **1.0380** ± 0.0038 |
| Barley $(-10^5)$ 5.0756 ± 0.0136 | 5.0697 ± **0.0039** | 5.0702 ± 0.0096 | **5.0696** ± **0.0039** | 5.0699 ± 0.0041 | 5.0699 ± 0.0041 |
| Diabetes $(-10^5)$ 1.9394 ± 0.0032 | 1.9391 ± 0.0034 | 1.9394 ± **0.0029** | **1.9386** ± 0.0034 | 1.9395 ± 0.0034 | 1.9393 ± 0.0035 |
| HailFinder $(-10^5)$ 4.9207 ± 0.0039 | 4.9206 ± 0.0038 | 4.9204 ± **0.0034** | 4.9202 ± 0.0040 | **4.9202** ± 0.0040 | 4.9207 ± 0.0037 |
| Mildew $(-10^5)$ 4.5426 ± 0.0096 | 4.5412 ± 0.0091 | 4.5417 ± 0.0101 | 4.5401 ± 0.0094 | **4.5388** ± **0.0083** | 4.5395 ± 0.0090 |
| Win95pts $(-10^4)$ 9.4322 ± 0.0448 | 9.4169 ± **0.0433** | **9.4086** ± 0.0452 | 9.4125 ± 0.0454 | 9.4154 ± 0.0457 | 9.4210 ± 0.0468 |

| $q_0$ | | | | | |
|---|---|---|---|---|---|
| 0.7 | 0.75 | 0.8 | 0.85 | 0.9 | 0.95 |
| Alarm $(-10^5)$ 1.0385 ± **0.0035** | 1.0388 ± **0.0035** | 1.0383 ± **0.0035** | **1.0382** ± **0.0035** | 1.0386 ± 0.0039 | 1.0383 ± 0.0040 |
| Barley $(-10^5)$ **5.0703** ± 0.0066 | 5.0704 ± 0.0065 | 5.0705 ± 0.0060 | 5.0708 ± **0.0057** | 5.0710 ± 0.0073 | 5.0720 ± 0.0126 |
| Diabetes $(-10^5)$ 1.9391 ± 0.0035 | 1.9391 ± 0.0033 | 1.9393 ± 0.0032 | 1.9393 ± 0.0035 | 1.9393 ± 0.0035 | **1.9390** ± **0.0030** |
| HailFinder $(-10^5)$ 4.9207 ± 0.0039 | 4.9208 ± **0.0035** | 4.9208 ± 0.0038 | 4.9205 ± 0.0038 | **4.9204** ± 0.0040 | **4.9204** ± 0.0040 |
| Mildew $(-10^5)$ **4.5362** ± **0.0069** | 4.5378 ± 0.0074 | 4.5389 ± 0.0084 | 4.5410 ± 0.0098 | 4.5430 ± 0.0097 | 4.5472 ± 0.0092 |
| Win95pts $(-10^4)$ 9.4200 ± 0.0453 | 9.4183 ± 0.0441 | 9.4199 ± 0.0462 | 9.4192 ± 0.0481 | 9.4160 ± 0.0468 | **9.4130** ± **0.0439** |

| $\beta$ | | | | | |
|---|---|---|---|---|---|
| 0.0 | 0.5 | 1.0 | 1.5 | 2.0 | 2.5 |
| Alarm $(-10^5)$ 1.0387 ± 0.0036 | **1.0378** ± 0.0038 | 1.0383 ± 0.0036 | 1.0386 ± 0.0035 | 1.0387 ± 0.0040 | 1.0387 ± **0.0034** |
| Barley $(-10^5)$ 5.0775 ± 0.0121 | 5.0694 ± 0.0043 | **5.0689** ± **0.0035** | 5.0696 ± 0.0040 | 5.0697 ± 0.0055 | 5.0698 ± 0.0097 |
| Diabetes $(-10^5)$ **1.9389** ± 0.0033 | 1.9390 ± 0.0034 | 1.9394 ± 0.0034 | 1.9393 ± **0.0032** | 1.9394 ± **0.0032** | 1.9391 ± 0.0035 |
| HailFinder $(-10^5)$ 4.9212 ± 0.0039 | 4.9205 ± **0.0037** | **4.9203** ± 0.0040 | 4.9206 ± 0.0038 | 4.9205 ± 0.0038 | 4.9205 ± 0.0040 |
| Mildew $(-10^5)$ 4.5378 ± **0.0075** | **4.5371** ± **0.0075** | 4.5392 ± 0.0090 | 4.5404 ± 0.0094 | 4.5440 ± 0.0102 | 4.5456 ± 0.0090 |
| Win95pts $(-10^4)$ 9.4515 ± 0.0505 | **9.4092** ± 0.0404 | 9.4135 ± 0.0420 | 9.4101 ± 0.0385 | 9.4116 ± 0.0444 | 9.4106 ± 0.0427 |

| $m$ | | | | | |
|---|---|---|---|---|---|
| 5 | 7 | 10 | 12 | 15 | 20 |
| Alarm $(-10^5)$ 1.0383 ± **0.0036** | 1.0386 ± **0.0036** | **1.0381** ± 0.0037 | 1.0389 ± 0.0037 | 1.0384 ± 0.0039 | 1.0384 ± **0.0036** |
| Barley $(-10^5)$ 5.0721 ± 0.0120 | 5.0707 ± 0.00061 | 5.0709 ± 0.0083 | **5.0704** ± 0.0066 | **5.0704** ± 0.0061 | 5.0705 ± **0.0060** |
| Diabetes $(-10^5)$ 1.9392 ± **0.0031** | 1.9392 ± 0.0036 | **1.9391** ± 0.0032 | 1.9392 ± 0.0032 | **1.9391** ± 0.0033 | 1.9394 ± 0.0034 |
| HailFinder $(-10^5)$ 4.9202 ± 0.0041 | 4.9209 ± 0.0037 | **4.9201** ± 0.0039 | 4.9210 ± 0.0040 | 4.9204 ± **0.0036** | 4.9210 ± 0.0037 |
| Mildew $(-10^5)$ 4.5440 ± 0.0097 | 4.5427 ± 0.0097 | 4.5409 ± 0.0093 | 4.5392 ± 0.0092 | 4.5386 ± **0.0082** | **4.5385** ± 0.0085 |
| Win95pts $(-10^4)$ 9.4201 ± **0.0452** | 9.4190 ± **0.0452** | 9.4178 ± 0.0457 | 9.4188 ± 0.0458 | 9.4164 ± 0.0462 | **9.4145** ± 0.0467 |

Table 10: Mean and standard deviation of the BDeu score for ACO-E for each parameter setting

| | $\rho$ | | | | | |
|---|---|---|---|---|---|---|
| | 0.0 | 0.1 | 0.2 | 0.3 | 0.4 | 0.5 |
| Alarm | $6.9 \pm 4.9$ | $5.6 \pm 3.1$ | $6.0 \pm 3.2$ | $5.6 \pm 3.3$ | $5.9 \pm 3.0$ | $\mathbf{5.5} \pm \mathbf{3.0}$ |
| Barley | $56.4 \pm 10.8$ | $52.8 \pm \mathbf{3.9}$ | $53.0 \pm 4.1$ | $53.2 \pm 4.4$ | $\mathbf{52.6} \pm 4.0$ | $52.9 \pm 4.8$ |
| Diabetes | $\mathbf{63.5} \pm 5.8$ | $65.5 \pm 5.3$ | $65.0 \pm 5.4$ | $65.1 \pm 5.6$ | $64.2 \pm 5.4$ | $63.7 \pm \mathbf{4.8}$ |
| HailFinder | $50.6 \pm 6.9$ | $\mathbf{50.5} \pm 6.8$ | $51.0 \pm 7.8$ | $51.8 \pm 7.8$ | $51.8 \pm \mathbf{6.7}$ | $51.1 \pm 7.9$ |
| Mildew | $25.7 \pm 5.8$ | $22.6 \pm 5.0$ | $22.8 \pm 5.1$ | $22.0 \pm 4.9$ | $\mathbf{21.4} \pm \mathbf{4.7}$ | $21.9 \pm 4.8$ |
| Win95pts | $94.6 \pm 27.7$ | $83.3 \pm 24.3$ | $\mathbf{81.7} \pm \mathbf{20.3}$ | $81.9 \pm 22.9$ | $81.9 \pm 24.2$ | $83.5 \pm 21.9$ |

| | $q_0$ | | | | | |
|---|---|---|---|---|---|---|
| | 0.7 | 0.75 | 0.8 | 0.85 | 0.9 | 0.95 |
| Alarm | $\mathbf{5.2} \pm \mathbf{2.5}$ | $5.6 \pm 3.1$ | $5.4 \pm 3.0$ | $6.1 \pm 3.7$ | $6.4 \pm 4.2$ | $6.7 \pm 4.1$ |
| Barley | $53.9 \pm 5.9$ | $53.6 \pm 5.8$ | $53.3 \pm \mathbf{5.4}$ | $53.4 \pm 5.9$ | $\mathbf{53.1} \pm 5.8$ | $53.6 \pm 7.2$ |
| Diabetes | $\mathbf{62.1} \pm 4.8$ | $62.6 \pm 4.9$ | $63.4 \pm 4.9$ | $64.7 \pm 5.4$ | $66.1 \pm 5.3$ | $68.1 \pm \mathbf{4.7}$ |
| HailFinder | $52.1 \pm 7.3$ | $51.9 \pm 7.5$ | $51.5 \pm 8.2$ | $50.7 \pm 7.1$ | $50.6 \pm 7.7$ | $\mathbf{50.0} \pm \mathbf{5.9}$ |
| Mildew | $\mathbf{20.2} \pm \mathbf{3.8}$ | $21.1 \pm 4.7$ | $21.5 \pm 4.8$ | $22.6 \pm 5.1$ | $24.3 \pm 5.4$ | $26.6 \pm 4.9$ |
| Win95pts | $90.3 \pm 23.5$ | $88.1 \pm 24.5$ | $84.9 \pm 22.6$ | $83.2 \pm 22.5$ | $81.0 \pm \mathbf{22.2}$ | $\mathbf{79.3} \pm 27.1$ |

| | $\beta$ | | | | | |
|---|---|---|---|---|---|---|
| | 0.0 | 0.5 | 1.0 | 1.5 | 2.0 | 2.5 |
| Alarm | $7.4 \pm 5.1$ | $\mathbf{4.3} \pm \mathbf{1.1}$ | $4.9 \pm 2.4$ | $5.4 \pm 2.8$ | $6.2 \pm 3.6$ | $7.2 \pm 3.7$ |
| Barley | $61.4 \pm 9.0$ | $52.0 \pm 3.9$ | $51.9 \pm \mathbf{3.0}$ | $\mathbf{51.7} \pm 3.3$ | $51.9 \pm 3.3$ | $52.0 \pm 3.7$ |
| Diabetes | $64.3 \pm \mathbf{4.9}$ | $64.3 \pm 5.5$ | $\mathbf{64.2} \pm 5.5$ | $64.7 \pm 5.1$ | $64.6 \pm \mathbf{5.9}$ | $64.8 \pm 5.5$ |
| HailFinder | $52.2 \pm \mathbf{7.0}$ | $52.0 \pm 6.5$ | $51.5 \pm 6.7$ | $\mathbf{50.1} \pm 8.0$ | $50.3 \pm 8.0$ | $50.5 \pm 7.5$ |
| Mildew | $\mathbf{20.9} \pm 4.9$ | $\mathbf{20.9} \pm \mathbf{4.0}$ | $21.9 \pm 5.0$ | $22.4 \pm 5.1$ | $24.6 \pm 5.4$ | $25.7 \pm 5.2$ |
| Win95pts | $109.1 \pm 28.6$ | $89.6 \pm 23.9$ | $79.8 \pm 17.9$ | $77.0 \pm 17.2$ | $77.1 \pm 19.3$ | $\mathbf{74.4} \pm \mathbf{15.5}$ |

| | $m$ | | | | | |
|---|---|---|---|---|---|---|
| | 5 | 7 | 10 | 12 | 15 | 20 |
| Alarm | $7.1 \pm 4.3$ | $6.6 \pm 3.8$ | $5.9 \pm 3.5$ | $5.2 \pm 2.8$ | $5.7 \pm 3.6$ | $\mathbf{5.1} \pm \mathbf{2.4}$ |
| Barley | $54.3 \pm 7.3$ | $\mathbf{52.8} \pm 5.3$ | $53.8 \pm 6.6$ | $53.2 \pm 5.7$ | $53.7 \pm 5.8$ | $53.1 \pm \mathbf{5.1}$ |
| Diabetes | $66.2 \pm 5.1$ | $65.8 \pm 5.9$ | $64.1 \pm 5.4$ | $64.5 \pm 5.6$ | $63.5 \pm 5.4$ | $\mathbf{62.8} \pm \mathbf{4.7}$ |
| HailFinder | $\mathbf{50.5} \pm 7.2$ | $\mathbf{50.5} \pm 6.7$ | $51.0 \pm \mathbf{6.5}$ | $51.8 \pm 7.0$ | $51.4 \pm 8.5$ | $51.6 \pm 7.9$ |
| Mildew | $24.6 \pm 5.3$ | $23.9 \pm 5.6$ | $22.7 \pm 5.5$ | $22.2 \pm 4.9$ | $21.6 \pm \mathbf{4.7}$ | $\mathbf{21.4} \pm \mathbf{4.7}$ |
| Win95pts | $83.3 \pm 23.3$ | $\mathbf{83.0} \pm \mathbf{21.9}$ | $85.6 \pm 26.2$ | $85.5 \pm 22.1$ | $85.5 \pm 25.0$ | $84.0 \pm 25.5$ |

Table 11: Mean and standard deviation of the SHD for ACO-E for each parameter setting

|  |  | ACO-E | GREEDY-E | ACO-B | EPQ |
|---|---|---|---|---|---|
| Alarm | SHD | **5.9 ± 3.5** | 21.9 ± 9.0 | 11.9 ± 11.9 | 26.1 ± 13.4 |
|  | Score ($-10^5$) | **1.0385 ± 0.0037** | 1.0389 ± 0.0039 | 1.0388 ± 0.0038 | 1.0415 ± 0.0045 |
| Barley | SHD | **53.5 ± 6.0** | 104.8 ± 9.7 | 67.3 ± 21.6 | 101.4 ± 14.4 |
|  | Score ($-10^5$) | **5.0708 ± 0.0078** | 5.2449 ± 0.0124 | 5.0944 ± 0.0423 | 5.2354 ± 0.0628 |
| Diabetes | SHD | **64.5 ± 5.4** | 69.2 ± **3.0** | 70.7 ± 8.5 | 77.2 ± 7.1 |
|  | Score ($-10^5$) | **1.9392 ± 0.0033** | 1.9394 ± **0.0033** | 1.9406 ± 0.0041 | 1.9457 ± 0.0048 |
| HailFinder | SHD | 51.1 ± 7.3 | **49.1 ± 0.8** | 74.1 ± 19.7 | 82.8 ± 18.2 |
|  | Score ($-10^5$) | **4.9206** ± 0.0038 | 4.9213 ± **0.0036** | 4.9248 ± 0.0058 | 4.9481 ± 0.0177 |
| Mildew | SHD | **22.7 ± 5.3** | 29.3 ± **0.7** | 36.1 ± 14.0 | 50.3 ± 13.8 |
|  | Score ($-10^5$) | **4.5407** ± 0.0093 | 4.5531 ± **0.0039** | 4.5548 ± 0.0170 | 4.6148 ± 0.0369 |
| Win95pts | SHD | **84.5 ± 24.1** | 104.9 ± **15.5** | 178.9 ± 58.8 | 220.1 ± 31.6 |
|  | Score ($-10^4$) | **9.4178 ± 0.0457** | 9.4649 ± 0.0466 | 9.4589 ± 0.0717 | 9.9181 ± 0.0970 |

Table 12: Mean and standard deviation for metaheuristic algorithms from Condition 1 results

|  | Alarm | Barley | HailFinder | Mildew |
|---|---|---|---|---|
| ACO-E | 16.4 ± 4.7 | **80.9 ± 5.3** | **55.0 ± 5.3** | **31.0 ± 3.6** |
| MMHC | **9.6 ± 7.0** | 102.6 ± 9.2 | 208.0 ± 1.6 | 58.4 ± 7.4 |
| OR1 $k = 5$ | 27.8 ± 10.0 | 109.6 ± 9.5 | 190.8 ± 14.1 | 70.6 ± 4.2 |
| OR1 $k = 10$ | 31.2 ± 11.1 | 113.6 ± 15.6 | 183.2 ± 14.9 | 75.6 ± 6.3 |
| OR1 $k = 20$ | 37.8 ± 9.4 | 136.4 ± 2.9 | 184.6 ± 17.2 | 75.0 ± 4.8 |
| OR2 $k = 5$ | 21.2 ± 4.6 | 120.0 ± 4.5 | 184.6 ± 14.5 | 69.2 ± 3.3 |
| OR2 $k = 10$ | 33.2 ± 5.4 | 109.2 ± 16.2 | 187.0 ± 15.7 | 64.0 ± 4.4 |
| OR2 $k = 20$ | 39.4 ± 6.5 | 116.8 ± 18.4 | 200.8 ± 9.2 | 67.4 ± 3.4 |
| SC $k = 5$ | 34.2 ± 3.6 | 129.6 ± 13.1 | 194.2 ± 2.5 | N/A |
| SC $k = 10$ | 20.4 ± 11.8 | N/A | N/A | N/A |
| GS | 58.8 ± 6.5 | 143.3 ± 7.3 | 204.2 ± 9.9 | 62.2 ± 12.2 |
| PC | 15.2 ± 1.5 | 610.0 ± 10.6 | 385.6 ± 12.5 | 421.2 ± 10.7 |
| TPDA | **9.6 ± 1.5** | 207.2 ± 4.0 | 255.4 ± 3.4 | 97.8 ± 6.8 |
| GES | N/A | 159.0 ± 0.0 | 154.6 ± 54.3 | 38.8 ± 0.8 |

Table 13: SHD mean and standard deviation for state-of-the-art algorithms

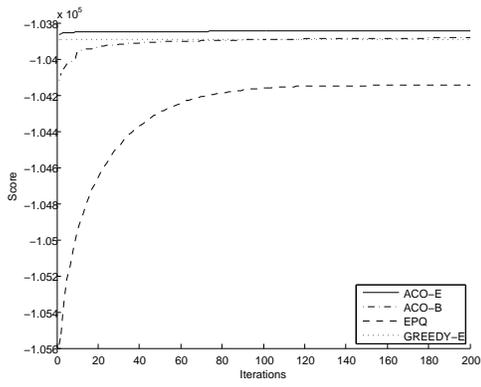

(a) Alarm

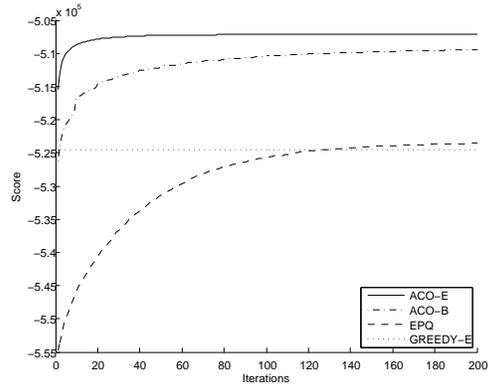

(b) Barley

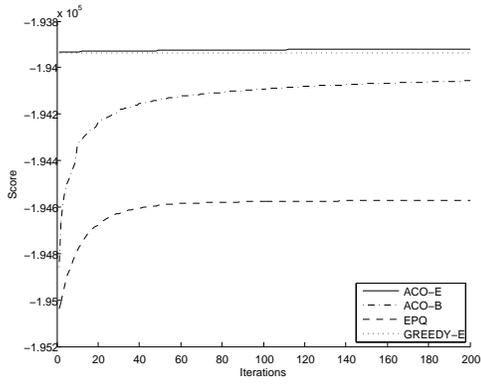

(c) Diabetes

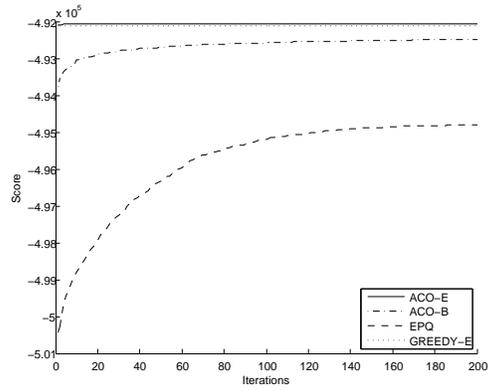

(d) HailFinder

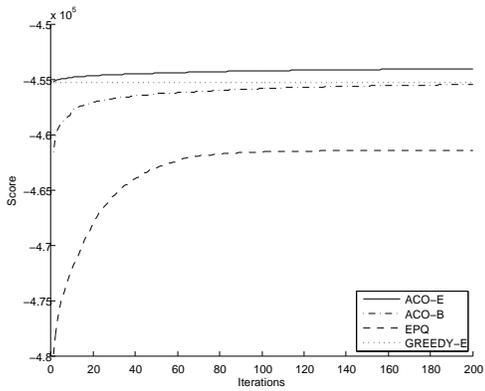

(e) Mildew

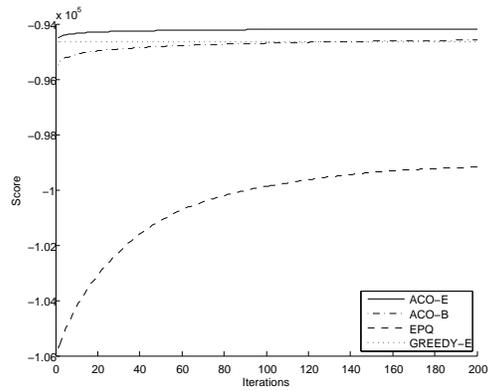

(f) Win95pts

Figure 9: Scores for metaheuristic algorithm comparison

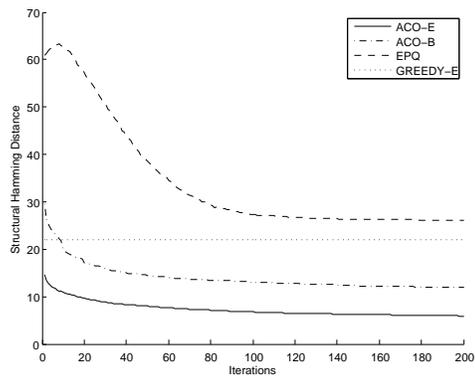

(a) Alarm

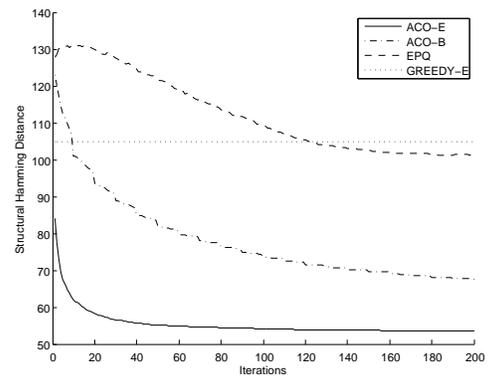

(b) Barley

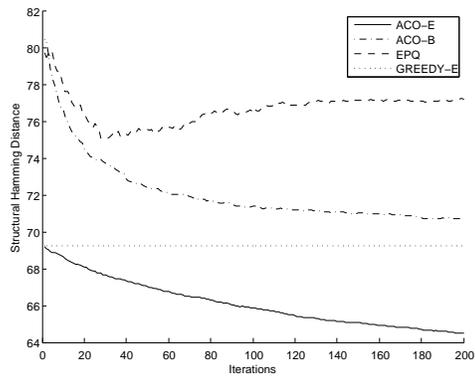

(c) Diabetes

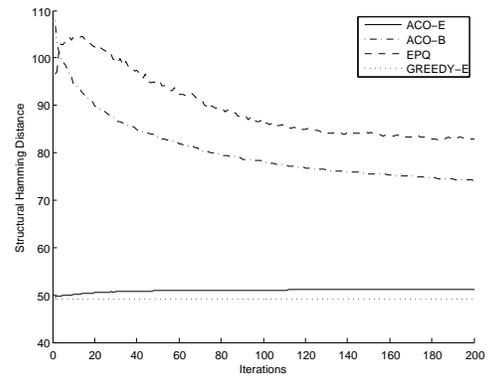

(d) HailFinder

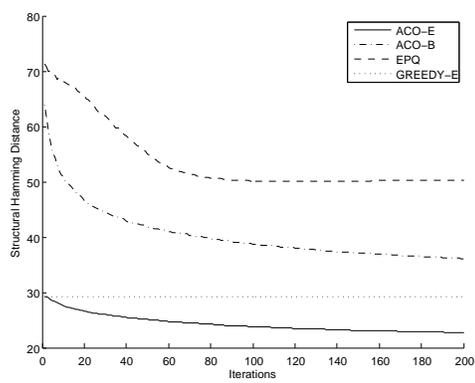

(e) Mildew

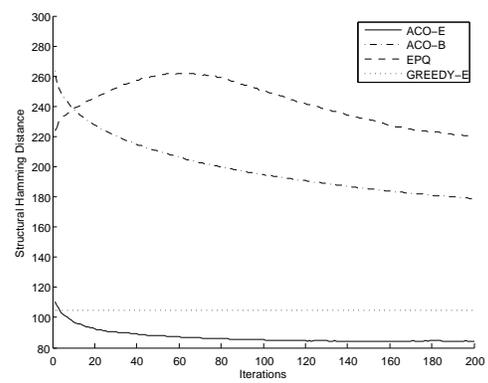

(f) Win95pts

Figure 10: SHD for metaheuristic algorithm comparison

| | | Network | | | | | |
|---|---|---|---|---|---|---|---|
| | | Alarm | Barley | Diabetes | HailFinder | Mildew | Win95pts |
| Sample Size | 100 | $49.32 \pm 8.37$ | $145.62 \pm 2.62$ | $78.24 \pm 5.28$ | $98.54 \pm 7.28$ | $84.49 \pm 1.34$ | $164.34 \pm 17.49$ |
| | 500 | $23.30 \pm 5.32$ | $132.25 \pm 5.32$ | $70.95 \pm 5.92$ | $79.99 \pm 11.60$ | $74.84 \pm 2.46$ | $91.45 \pm 21.75$ |
| | 1000 | $17.73 \pm 4.58$ | $106.05 \pm 3.85$ | $68.13 \pm 7.65$ | $69.61 \pm 8.00$ | $55.53 \pm 2.78$ | $77.10 \pm 15.77$ |
| | 5000 | $6.45 \pm 2.71$ | $66.30 \pm 5.15$ | $67.15 \pm 4.30$ | $58.50 \pm 5.99$ | $36.68 \pm 5.21$ | $56.29 \pm 14.53$ |
| | 10000 | $4.33 \pm 1.74$ | $51.49 \pm 2.82$ | $61.01 \pm 3.18$ | $52.64 \pm 6.85$ | $18.96 \pm 0.79$ | $50.84 \pm 11.19$ |

Table 14: Structural Hamming distance for different sample sizes

| | | Network | | | | | |
|---|---|---|---|---|---|---|---|
| | | Alarm $(-10^5)$ | Barley $(-10^5)$ | Diabetes $(-10^5)$ | HailFinder $(-10^5)$ | Mildew $(-10^5)$ | Win95pts $(-10^4)$ |
| Sample Size | 100 | $0.0138 \pm 0.0005$ | $0.0774 \pm 0.0006$ | $0.0316 \pm 0.0009$ | $0.0596 \pm 0.0003$ | $0.0667 \pm 0.0006$ | $0.1507 \pm 0.0006$ |
| | 500 | $0.0568 \pm 0.0010$ | $0.3263 \pm 0.0041$ | $0.1135 \pm 0.0028$ | $0.2687 \pm 0.0012$ | $0.2946 \pm 0.0019$ | $0.5446 \pm 0.0012$ |
| | 1000 | $0.1092 \pm 0.0015$ | $0.5833 \pm 0.0028$ | $0.2102 \pm 0.0013$ | $0.5189 \pm 0.0023$ | $0.5576 \pm 0.0022$ | $1.0198 \pm 0.0016$ |
| | 5000 | $0.5228 \pm 0.0029$ | $2.6028 \pm 0.0024$ | $0.9810 \pm 0.0027$ | $2.4807 \pm 0.0027$ | $2.4178 \pm 0.0037$ | $4.7468 \pm 0.0035$ |
| | 10000 | $1.0388 \pm 0.0037$ | $5.0695 \pm 0.0035$ | $1.9399 \pm 0.0032$ | $4.9205 \pm 0.0037$ | $4.5338 \pm 0.0043$ | $9.3794 \pm 0.0043$ |

Table 15: Training score for different sample sizes

| | | Network | | | | | |
|---|---|---|---|---|---|---|---|
| | | Alarm $(-10^5)$ | Barley $(-10^5)$ | Diabetes $(-10^5)$ | HailFinder $(-10^5)$ | Mildew $(-10^5)$ | Win95pts $(-10^4)$ |
| Sample Size | 100 | $0.0143 \pm 0.0005$ | $0.0778 \pm 0.0007$ | $0.0316 \pm 0.0007$ | $0.0604 \pm 0.0004$ | $0.0668 \pm 0.0006$ | $0.1669 \pm 0.0007$ |
| | 500 | $0.0571 \pm 0.0009$ | $0.3272 \pm 0.0042$ | $0.1136 \pm 0.0028$ | $0.2690 \pm 0.0012$ | $0.2947 \pm 0.0015$ | $0.5564 \pm 0.0011$ |
| | 1000 | $0.1092 \pm 0.0014$ | $0.5835 \pm 0.0026$ | $0.2104 \pm 0.0012$ | $0.5192 \pm 0.0019$ | $0.5581 \pm 0.0021$ | $1.0238 \pm 0.0014$ |
| | 5000 | $0.5229 \pm 0.0027$ | $2.6031 \pm 0.0030$ | $0.9811 \pm 0.0025$ | $2.4811 \pm 0.0030$ | $2.4181 \pm 0.0038$ | $4.7535 \pm 0.0031$ |
| | 10000 | $1.0387 \pm 0.0039$ | $5.0702 \pm 0.0030$ | $1.9394 \pm 0.0037$ | $4.9214 \pm 0.0036$ | $4.5350 \pm 0.0045$ | $9.3883 \pm 0.0040$ |

Table 16: Test score for different sample sizes

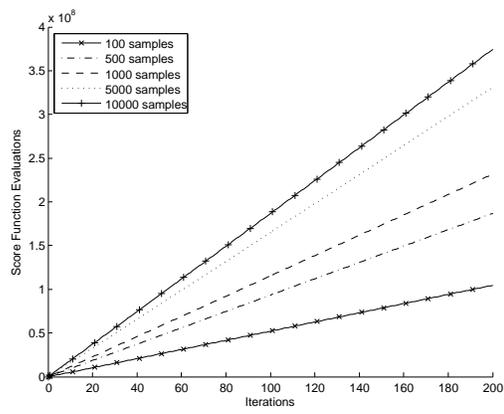

(a) Alarm

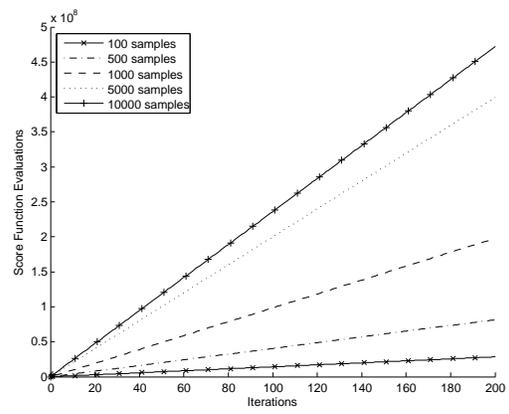

(b) Barley

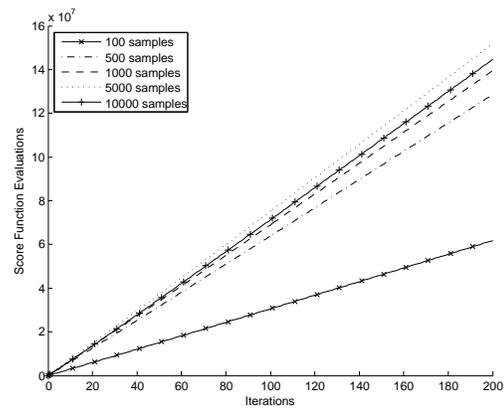

(c) Diabetes

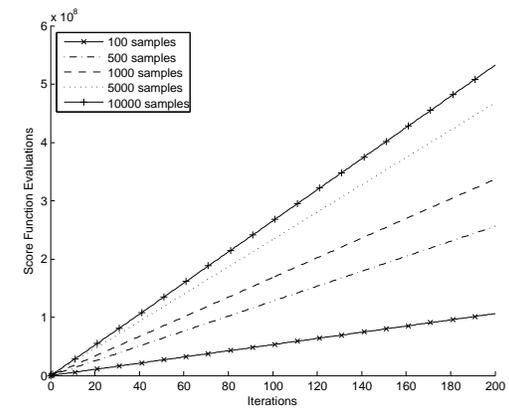

(d) HailFinder

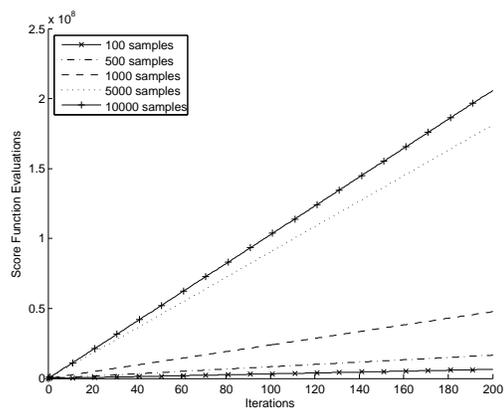

(e) Mildew

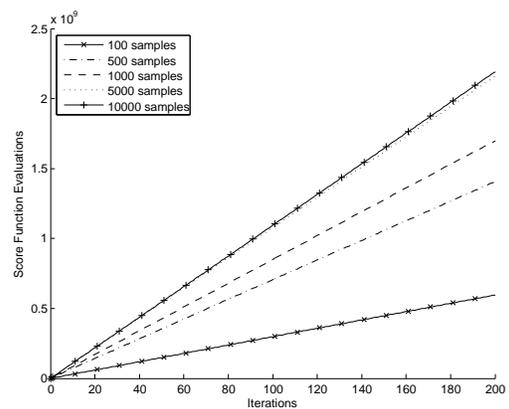

(f) Win95pts

Figure 11: Score function evaluations for different sample sizes

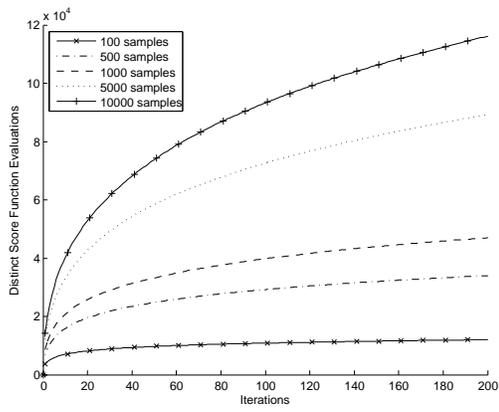

(a) Alarm

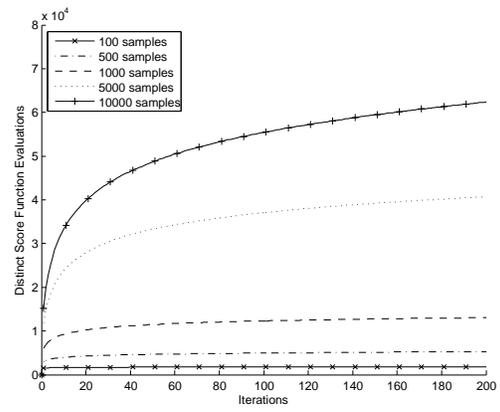

(b) Barley

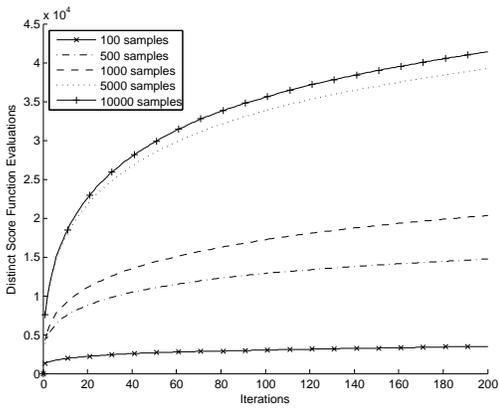

(c) Diabetes

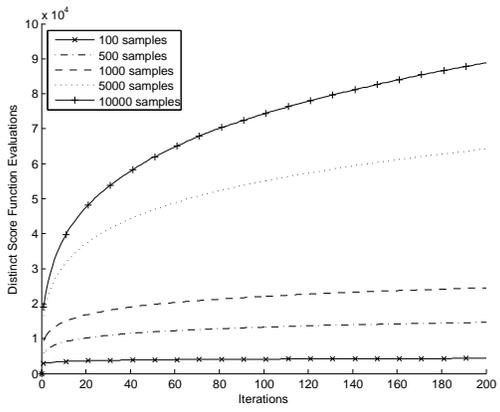

(d) HailFinder

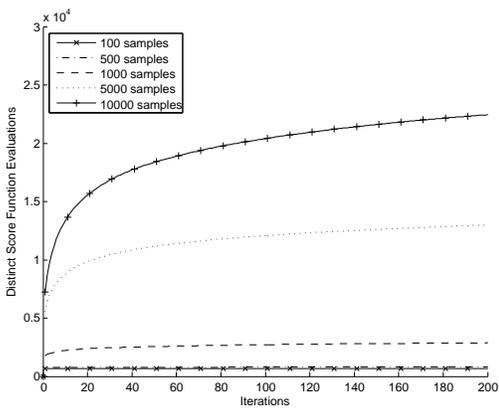

(e) Mildew

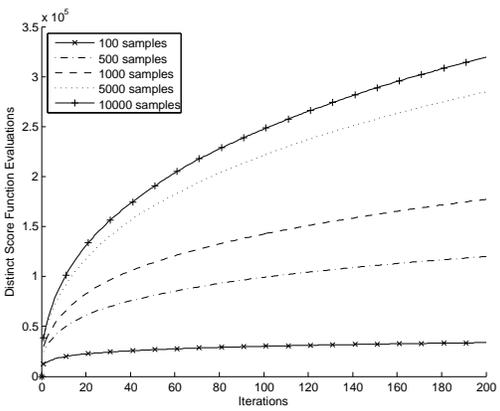

(f) Win95pts

Figure 12: Distinct score function evaluations for different sample sizes

| | | ACO-E | GREEDY-E | ACO-B | EPQ |
|---|---|---|---|---|---|
| Alarm | SHD | **4.33** ± 1.74 | 24.17 ± 9.16 | 5.98 ± 4.63 | 16.09 ± 9.92 |
| | Score ($-10^5$) | **1.0388** ± 0.0037 | 1.0396 ± 0.0038 | 1.0388 ± 0.0040 | 1.0389 ± 0.0037 |
| | Test Score ($-10^5$) | **1.0387** ± 0.0039 | 1.0395 ± 0.0044 | 1.0391 ± 0.0038 | 1.0396 ± 0.0037 |
| | Score Eval. | 3.7e8 ± 2.7e7 | **6.8e4** ± 7.3e3 | 1.7e7 ± 1.9e5 | 3.2e7 ± 1.3e6 |
| | Dist. Score Eval. | 1.2e5 ± 5.6e3 | **2.9e3** ± 2.1e2 | 7.2e4 ± 2.3e3 | 2.7e4 ± 2.3e3 |
| Barley | SHD | **51.49** ± 2.82 | 106.58 ± 8.95 | 52.95 ± 3.71 | 91.18 ± 17.36 |
| | Score ($-10^5$) | **5.0695** ± 0.0035 | 5.2415 ± 0.0114 | 5.0698 ± 0.0033 | 5.1677 ± 0.0740 |
| | Test Score ($-10^5$) | **5.0702** ± 0.0030 | 5.2413 ± 0.0116 | 5.0702 ± 0.0034 | 5.1673 ± 0.0734 |
| | Score Eval. | 4.7e8 ± 1.5e7 | **1.3e5** ± 5.7e3 | 3.1e7 ± 1.5e5 | 4.5e7 ± 2.1e6 |
| | Dist. Score Eval. | 6.2e4 ± 1.3e3 | **4.1e3** ± 1.3e2 | 5.8e4 ± 1.0e3 | 4.3e4 ± 2.7e3 |
| Diabetes | SHD | **61.01** ± 3.18 | 68.71 ± 3.13 | 66.97 ± 4.88 | 77.13 ± 7.10 |
| | Score ($-10^5$) | 1.9399 ± 0.0032 | 1.9397 ± 0.0033 | **1.9392** ± 0.0039 | 1.9451 ± 0.0047 |
| | Test Score ($-10^5$) | **1.9394** ± 0.0037 | 1.9395 ± 0.0030 | 1.9398 ± 0.0032 | 1.9449 ± 0.0044 |
| | Score Eval. | 1.4e8 ± 7.5e6 | **2.7e4** ± 1.6e3 | 1.6e7 ± 1.0e5 | 2.7e7 ± 2.7e6 |
| | Dist. Score Eval. | 4.1e4 ± 1.8e3 | **2.2e3** ± 3.1e1 | 3.3e4 ± 1.0e3 | 1.6e4 ± 1.6e3 |
| HailFinder | SHD | 52.64 ± 6.85 | **49.20** ± 0.89 | 61.59 ± 11.63 | 78.81 ± 16.32 |
| | Score ($-10^5$) | 4.9205 ± 0.0037 | 4.9212 ± 0.0039 | 4.9213 ± 0.0036 | 4.9293 ± 0.0073 |
| | Test Score ($-10^5$) | 4.9214 ± 0.0036 | **4.9209** ± 0.0038 | 4.9214 ± 0.0038 | 4.9296 ± 0.0080 |
| | Score Eval. | 5.3e8 ± 3.3e7 | **1.0e5** ± 2.7e3 | 4.0e7 ± 2.9e5 | 6.1e7 ± 3.7e6 |
| | Dist. Score Eval. | 8.9e4 ± 3.2e3 | **5.4e3** ± 8.7e1 | 6.9e4 ± 2.2e3 | 5.5e4 ± 3.9e3 |
| Mildew | SHD | **18.96** ± 0.79 | 29.22 ± 0.77 | 19.41 ± 3.83 | 43.59 ± 11.79 |
| | Score ($-10^5$) | **4.5338** ± 0.0043 | 4.5527 ± 0.0038 | 4.5348 ± 0.0058 | 4.5982 ± 0.0292 |
| | Test Score ($-10^5$) | **4.5350** ± 0.0045 | 4.5526 ± 0.0044 | 4.5350 ± 0.0052 | 4.5989 ± 0.0299 |
| | Score Eval. | 2.1e8 ± 9.8e6 | **4.2e4** ± 1.1e3 | 1.6e7 ± 1.6e5 | 2.8e7 ± 2.4e6 |
| | Dist. Score Eval. | 2.2e4 ± 5.4e2 | **2.2e3** ± 3.7e1 | 1.5e4 ± 2.6e2 | 1.5e4 ± 1.2e3 |
| Win95pts | SHD | **50.84** ± 11.19 | 85.75 ± 16.44 | 91.08 ± 18.52 | 231.25 ± 42.46 |
| | Score ($-10^4$) | **9.3794** ± 0.0043 | 9.4121 ± 0.0043 | 9.3890 ± 0.0036 | 9.6061 ± 0.0075 |
| | Test Score ($-10^4$) | **9.3883** ± 0.0040 | 9.4153 ± 0.0045 | 9.3897 ± 0.0042 | 9.6058 ± 0.0080 |
| | Score Eval. | 2.2e9 ± 2.4e8 | **5.2e5** ± 8.9e4 | 8.5e7 ± 8.6e5 | 1.2e8 ± 4.6e6 |
| | Dist. Score Eval. | 3.2e5 ± 1.9e4 | **1.5e4** ± 7.9e2 | 2.7e5 ± 1.2e4 | 1.9e5 ± 5.4e3 |

Table 17: Mean and standard deviation for tuned metaheuristic algorithms

| | | Alarm | Barley | Diabetes | HailFinder | Mildew | Win95pts |
|---|---|---|---|---|---|---|---|
| Sample Size | 100 | 0.25 | 0.01 | 0.06 | 0.06 | 0.00 | 0.53 |
| | 500 | 0.40 | 0.12 | 0.32 | 0.18 | 0.02 | 1.30 |
| | 1000 | 0.46 | 0.32 | 0.34 | 0.28 | 0.12 | 1.53 |
| | 5000 | 0.57 | 0.72 | 0.36 | 0.44 | 0.71 | 1.96 |
| | 10000 | 0.58 | 0.93 | 0.30 | 0.45 | 0.82 | 2.11 |

Table 18: Mean number of v-structures divided by number of nodes on greedy searches

## 8. Discussion

This section will discuss the results presented in the previous section. In general, the discussion will involve looking at the score and SHD values (as defined in Section 6.2.2) obtained by the algorithms. It should be noted that a better score does not necessarily mean a better SHD value and vice-versa. This can occur because of small sample sizes and because of the parameters given to the scoring function (such as the equivalent sample size and $\kappa$ value), which have been shown to produce differences in scoring function behavior (Kayaalp & Cooper, 2002). In general, different data sets have different parameter values at which they behave optimally. There does not seem to be a general method to find the optimum values. This problem has been looked at in some depth by Silander, Kontkanen et al. (2007).

The first figures to be examined will be those in Tables 10 and 11 from Condition 1. These presented the results of experiments that varied the parameter values of the ACO-E algorithm. Looking at these figures, there is evidence that the ACO-E algorithm provides useful behavior for reasonable values of the parameters.

Next, the results from experimental Condition 5 will be examined, particularly Table 18 in the context of the Bayesian network properties given in Table 5. Along with the results which show behavior of ACO-E as a function of sample size in Condition 3 (Tables 14, 15 and 16), this discussion will seek to characterize ACO-E performance from the perspective of the generating network and sample. Evidence will be presented that shows ACO-E performs better with more complicated networks, i.e. networks with more v-structures.

The previous discussion focuses on the behavior of ACO-E as a function of its various parameters. The next results that will be looked at are intended to provide a comparison against other Bayesian network structure learning algorithms. These include Figures 9 and 10 and Table 12 from Condition 1 and Table 17 from Condition 4. These present ACO-E against other metaheuristic algorithms that are similar. In these results there is strong evidence that ACO-E is performing well against the other algorithms.

Also from a comparative perspective, the results given in Table 13 will be discussed. These present a series of tests comparing ACO-E to other state-of-the-art Bayesian network structure learning algorithms. Again, looking at the figures, there is strong evidence that ACO-E is competitive in its performance.

Finally, the complexity results from Condition 3 will be shown in the form of Figures 11 and 12.

In order to perform a comparison, statistical tests will be needed. Because of the non-normality of the distributions of some of the results, tests others than ones which rely on the normality of the data will be used. These are mentioned below.

### 8.1 ACO-E Behavior

In this section the behavior of ACO-E as its parameters are varied will be analyzed. As shown in Tables 10 and 11 there is evidence that there is a difference in the behavior of the ACO-E algorithm depending on the input parameters. These differences will be analyzed using the two-tailed Mann-Whitney $U$ test or Student's $T$ test. The particular test used depends on the normality of the data, which can be tested with the Jarque-Bera test.

In order to perform this comparison, the best figures from Tables 10 to 11 will be compared to the situation where that particular part of the ACO-E algorithm has been turned off. E.g. in Table 10 on the Alarm row, the best figure is at $\rho = 0.5$. This is compared to the value at $\rho = 0.0$, as at this value no pheromone deposition or evaporation is occurring. The values at which the various parts of the ACO-E algorithm have been 'turned off' are $\rho = 0.0$, $q_0 = 1$ and $\beta = 0.0$. For the value of $q_0 = 1$, the algorithm behaves purely in a greedy fashion. Therefore for the purposes of testing, the value of the GREEDY-E algorithm in Table 12 will be used for comparison, as these results would be exactly the same as the case where $q_0 = 1$. The results of these comparisons are shown in Table 19. This table shows p-values for each comparison.

### 8.1.1 THE BEHAVIOR OF $\rho$

Looking at Table 19 the results for $\rho$ that seem most certain are those for Barley, Mildew and Win95pts. Looking at Tables 10 and 11 for these networks, the values of $\rho$ are in the $0.2 - 0.4$ range. Also looking at the features of these networks in Table 5 there is a correspondence of $\rho = 0.2$ to 76 nodes (Win95pts), $\rho = 0.3$ to 48 nodes (Barley) and $\rho = 0.4$ to 35 nodes (Mildew). Whilst not conclusive, this suggests that $\rho$ behaves well in the region $0.2 - 0.4$ (for those data sets that it works at all). The fact that there is not much variance in this range for these networks means this range is quite robust. There also is a suggestion that datasets with more nodes would use smaller values of $\rho$. This makes sense, as larger networks would probably need to spend more time following the best solutions, as a low value of $\rho$ would provide.

### 8.1.2 THE BEHAVIOR OF $q_0$

The parameter $q_0$ appears to have an effect on most of the networks, with the possible exception of HailFinder. For some of the networks (Alarm and Barley) the parameter has a large effect over a wide range, whereas for others (Diabetes, Mildew and Win95pts), the effect depends to a large extent on the value for $q_0$. The largest effects from a scoring function point of view appear to be on the Barley, Mildew and Win95pts networks.

Looking at these networks, the large variations in behavior across different values of $q_0$ make it difficult to predict what the best value of the parameter might be for a particular data set. One rule of thumb might be that smaller values of $q_0$ create more exploration and so might be useful for smaller data sets, whereas larger data sets need more exploitation in order to get to a reasonable answer.

### 8.1.3 THE BEHAVIOR OF $\beta$

From Table 19, the networks for which the parameter $\beta$ plays the most role appear to be Alarm, Barley and Win95pts. Because of the differences of the best values between the scoring function and the SHD it is difficult to predict the best value for $\beta$. In the case of Barley, the behavior is quite robust to values of $\beta$ in the range $0.5 - 2.5$. However, for Alarm and Win95pts, the behavior depends on the value of the parameter with a smaller value being better for Alarm and a larger value for Win95pts. As a rule of thumb it appears that networks with less numbers of nodes need smaller values of $\beta$ to help avoid local minima, whereas networks with more nodes need larger values of $\beta$ in order to focus the search more effectively.

### 8.1.4 THE BEHAVIOR OF $m$

Looking at Tables 10 and 11 it can be seen that the value of $m$ can sometimes have a small effect on the effectiveness of ACO-E. In this case, the effect is most pronounced on the Alarm, Diabetes and Mildew networks, with higher values of $m$ giving a smaller SHD. Indeed in all cases, higher values of $m$ never produce statistically worse results, as is to be expected. However, it is important to bear in mind the increased running times with larger values of $m$.

### 8.1.5 GENERAL DISCUSSION

The reason for the strange behavior of the HailFinder results can possibly be explained by examining it's graphs of score function and SHD against time (Figures 9 and 10). It can be seen that as the score is improving over iterations, the SHD value is *deteriorating*. This might lead one to the conclusion that there is a problem with the scoring function for the HailFinder case, perhaps with its parameters. Another plausible reason for the HailFinder and Win95pts results being out of sync with the others is that they are larger networks, which might favour more aggressive exploitation of the best-so-far solution than the smaller ones. In this case, this would correspond to lower values of $\rho$ and higher values of $q_0$. Also heuristic information might be more useful with large numbers of variables, leading to the better results with large values of $\beta$. Note that these problems with the HailFinder network have also been seen by de Campos and Castellano (2007).

## 8.2 Behavior of ACO-E with Respect to Test Network and Sample

In the previous section, it was seen that ACO-E can be a useful algorithm in learning the structure of Bayesian networks. It was also seen that the values of the parameters that produced the best behavior depended on the network that was being tested. Some rules of thumb that consolidate the characteristics observed in the previous section were:

- For data with more variables, have lower values of $\rho$, higher values of $q_0$ and higher values of $\beta$.

- For data with less variables, have higher values of $\rho$, lower values of $q_0$ and lower values of $\beta$.

However, it was also seen that ACO-E is not always very successful in learning. This was because little difference was seen when certain parameters were 'turned off' with certain networks. Looking again at Table 19, it seems that the networks for which the effect was most felt were the Barley, Mildew and Win95pts networks. But why is this?

Looking at Table 5 there does not seem to be any discernible pattern between the network properties and the suitability of the algorithm. However, the values of the number of v-structures normalised by the number of nodes in the graph show a more definite reason. The networks with which ACO-E performed well all have a larger number of nodes that have a higher in-degree and hence a larger number of v-structures. As a result of this, data sampled from these networks is better going to match a similar network in the scoring function, i.e. one that is similar to the standard network. Because the search starts from the empty graph, it is more likely that the search would get trapped in a local minimum in trying to add enough arcs to get to the needed number. Due to ACO-E being a stochastic algorithm, it is able to avoid these local minima.

The upshot of this is that ACO-E would be a good candidate algorithm for data sampled from networks with a large number of v-structures. However, in the real world a generating network does not exist. Therefore, the experiments of Condition 5 were designed to try and estimate this quantity. The results of these experiments are shown in Table 18. It can be seen that there is some association between the results when the number of samples is at 10000 and the average number of v-structures per node. Indeed, the value of the correlation coefficient between the values is $r = 0.94$, which indicates a linear relationship with a p-value of 0.006. However, the results are not quite the same when the number of samples decrease. For example, at 100 samples the correlation is not visible. This makes sense, as the low number of samples would not be able to support many v-structures. As such, the estimate might only be valid in the large sample limit.

However, the procedure employed in Condition 5 does indicate a way in seeing how effective of the ACO-E algorithm would be on arbitrary set of data. If the value calculated by the resampling method was low (towards 0), then ACO-E would probably not be particularly effective and a simpler algorithm would perform well. However, as the value rises, the probable number of v-structures also rises and hence ACO-E (and other methods designed to avoid local maxima) would fare better.

These ideas seem to be borne out by examining Table 14. It appears that high expected values of v-structures per node imply good performance of the ACO-E algorithm, i.e. there is an obvious large improvement in SHD. On the contrary, low expected values of v-structures per node are associated with small improvements in the SHD as the algorithm progresses.

## 8.3 Metaheuristic Algorithm Comparison

Figures 9 and 10 and Table 12 show the results of comparing ACO-E against other metaheuristic algorithms. It can be seen that ACO-E performs better than the other algorithms shown except in the case of the HailFinder network, where GREEDY-E gives a better result for the SHD. However, in this case, ACO-E gives a better score value. This is the same as the problem discussed in Section 8.1, that gave a better score for a worse structure.

These statements can be backed up by looking at Table 20 which gives p-values for a two-tailed unpaired Mann-Whitney-U test comparing ACO-E results against the other algorithms after runs had ended. With this statistic, the smaller the number, the more significant the test. Since the results from each of the runs comes from a separate sample of the network, the correct tests would be unpaired. The data used in the tests were those from the metaheuristic algorithm comparison, i.e. over all combinations of the parameters for ACO-E and ACO-B. It seems that in most cases, the results are highly significant, which supports the assumption that ACO-E performs well. In the cases where the significance is not so high (ACO-E score compared to GREEDY-E score with the Alarm network and ACO-E score compared to GREEDY-E score with the Diabetes network), it should be noted that tiny changes to the score value can lead to large structural changes as an algorithm converges towards the optimum (generating) network. In these cases, the SHD p-values show a highly significant difference.

Comparisons were also made between the variances of the results as seen in Table 21, which gives p-values for Conover's (1999) Squared Ranks one-tailed test. From this table can be seen that ACO-E generally has a lower standard deviation in its results after finishing its run compared to ACO-B and EPQ. Whilst the standard deviation of results compared to GREEDY-E are significantly lower with respect to the Alarm and Barley networks, in the other cases GREEDY-E seems to be the most consistent with regard to its final results.

It should be noted that non-parametric tests were used, as the results in general had non-normal distributions. It should also be noted that some of the results in the tables might seem incorrect. E.g. in Table 20, in the Win95pts-Score row, the test for ACO-B is more significant than that for GREEDY-E, even though the mean of GREEDY-E is further from ACO-E than that of ACO-B in Table 12. This is because of the larger sample size for the ACO-B test, which had 1296 samples, compared to the 216 samples for GREEDY-E.

|  |  | $\rho$ | $q_0$ | $\beta$ |
|---|---|---|---|---|
| Alarm | SHD | $8.0 \times 10^{-4}$ | $4.3 \times 10^{-91}$ | $1.3 \times 10^{-16}$ |
|  | Score | $1.7 \times 10^{-1}$ | $2.1 \times 10^{-2}$ | $9.1 \times 10^{-3}$ |
| Barley | SHD | $1.3 \times 10^{-6}$ | $4.2 \times 10^{-230}$ | $6.9 \times 10^{-41}$ |
|  | Score | $1.3 \times 10^{-9}$ | $2.9 \times 10^{-72}$ | $2.5 \times 10^{-26}$ |
| Diabetes | SHD | $1.0 \times 10^{0}$ | $3.6 \times 10^{-42}$ | $8.7 \times 10^{-1}$ |
|  | Score | $9.2 \times 10^{-3}$ | $2.8 \times 10^{-1}$ | $1.0 \times 10^{0}$ |
| HailFinder | SHD | $9.3 \times 10^{-1}$ | *$3.3 \times 10^{-2}$* | $5.5 \times 10^{-3}$ |
|  | Score | $1.9 \times 10^{-1}$ | $1.1 \times 10^{-2}$ | $2.7 \times 10^{-2}$ |
| Mildew | SHD | $5.1 \times 10^{-16}$ | $3.6 \times 10^{-125}$ | $1.0 \times 10^{0}$ |
|  | Score | $1.4 \times 10^{-5}$ | $8.0 \times 10^{-63}$ | $3.5 \times 10^{-1}$ |
| Win95pts | SHD | $4.9 \times 10^{-8}$ | $9.0 \times 10^{-41}$ | $5.3 \times 10^{-44}$ |
|  | Score | $1.3 \times 10^{-7}$ | $2.0 \times 10^{-26}$ | $2.2 \times 10^{-18}$ |

Table 19: Comparisons of parameter behavior

|  |  | GREEDY-E | ACO-B | EPQ |
|---|---|---|---|---|
| Alarm | SHD | $1.1 \times 10^{-113}$ | $1.6 \times 10^{-31}$ | $1.9 \times 10^{-118}$ |
|  | Score | $4.3 \times 10^{-2}$ | $6.0 \times 10^{-3}$ | $1.8 \times 10^{-18}$ |
| Barley | SHD | $9.6 \times 10^{-124}$ | $1.8 \times 10^{-92}$ | $9.3 \times 10^{-123}$ |
|  | Score | $5.0 \times 10^{-123}$ | $5.9 \times 10^{-96}$ | $7.0 \times 10^{-123}$ |
| Diabetes | SHD | $3.1 \times 10^{-34}$ | $4.7 \times 10^{-104}$ | $1.5 \times 10^{-88}$ |
|  | Score | $2.5 \times 10^{-1}$ | $2.3 \times 10^{-18}$ | $4.5 \times 10^{-69}$ |
| HailFinder | SHD | $3.8 \times 10^{-12}$ | $2.7 \times 10^{-237}$ | $1.8 \times 10^{-99}$ |
|  | Score | $4.4 \times 10^{-3}$ | $2.2 \times 10^{-93}$ | $4.6 \times 10^{-113}$ |
| Mildew | SHD | $2.3 \times 10^{-50}$ | $4.7 \times 10^{-160}$ | $7.5 \times 10^{-115}$ |
|  | Score | $7.4 \times 10^{-62}$ | $5.7 \times 10^{-114}$ | $5.3 \times 10^{-118}$ |
| Win95pts | SHD | $1.5 \times 10^{-42}$ | $0$ | $4.8 \times 10^{-123}$ |
|  | Score | $3.0 \times 10^{-37}$ | $4.1 \times 10^{-53}$ | $5.0 \times 10^{-123}$ |

Table 20: p-values for Mann-Whitney $U$ test, 10,000 samples

Further results from Condition 4 confirm these findings. In those experiments, the same algorithms were run with the parameters tuned. The results of these experiments can be seen in Table 17. The findings from these results are similar to the ones discussed above, with some differences. ACO-E outperforms the other algorithms with the SHD and test score measures in all cases except for the HailFinder network, where GREEDY-E is better, as above. The differences between ACO-E and ACO-B, its main competitor, are not as pronounced, but still exist. With the Alarm, Barley and Mildew networks the practical difference is quite small, whereas with the Diabetes, HailFinder and Win95pts networks it is still quite large. However, even with this, ACO-E can be said to perform better in three areas:

- ACO-E is more robust with respect to the parameter values input. Comparing the tuned-parameter-value results to the results across all parameter values, it can be seen that ACO-E is not as sensitive to the values as ACO-B. This implies that ACO-E could be used in a learning problem without a long parameter optimization stage. Note that reasonable parameter values are still important, as discussed in Section 8.1.

- ACO-E converges faster to optimum values than ACO-B, in terms of the number of iterations. For example, in the Barley, Mildew and Win95pts cases, ACO-E reaches it's best SHD value in 20 iterations, whereas ACO-B takes about 200 iterations.

- ACO-E generally provides a smaller variance in output values than the other algorithms. This can be important in situations where a robust output is needed.

### 8.4 State-of-the-Art Algorithm Comparison

In this section, the comparison of ACO-E against other state-of-the-art Bayesian network structure learning algorithms will be analyzed. As shown in Table 13, ACO-E appears to have good performance against these other algorithms. The results of statistical comparisons of ACO-E against these algorithms are shown in Table 22.

In this table are shown p-values for individual comparisons of ACO-E against the other algorithms. The test used for all these comparisons was the Mann-Whitney $U$ test. This test was used, as the distributions were found to be not normal. At the foot of the table is the combined p-value found from the individual p-values above it. This is the total p-value for comparing ACO-E against all the other algorithms. The method of combining these values was

$$p_{combined} = 1 - \prod_{i=1}^{n} 1 - p_i,$$

where $p_i$ is the p-value of entry $i$ in the table, there being $n$ values in total. This method of combining the p-values is needed because of the chance of causing a Type I error otherwise. A Type I error is a false positive result, i.e. the null hypothesis is rejected when it should not be. This can occur in this case because if an experiment with a small chance of failing is repeated enough times, there will be a large chance that at least one of them will fail. It should be noted that the value at the foot of Alarm does not combine all the p-values above it. Instead it leaves out those of 'SC $k = 10$', 'PC' and 'OR2 $k = 5$'. This was because the median results for these figures were close to that of ACO-E and would have pushed the p-value very high. Therefore, the overall test is only valid for the tests that do not include the three just mentioned.

|  |  | GREEDY-E | ACO-B | EPQ |
|---|---|---|---|---|
| Alarm | SHD | $3.6 \times 10^{-84}$ | $2.4 \times 10^{-252}$ | $5.3 \times 10^{-104}$ |
|  | ESHD | $4.1 \times 10^{-103}$ | 0 | $7.3 \times 10^{-117}$ |
|  | Score | $8.0 \times 10^{-2}$ | $1.3 \times 10^{-1}$ | $5.0 \times 10^{-5}$ |
| Barley | SHD | $1.3 \times 10^{-61}$ | $1.5 \times 10^{-274}$ | $2.3 \times 10^{-66}$ |
|  | ESHD | $6.2 \times 10^{-44}$ | $7.6 \times 10^{-277}$ | $1.6 \times 10^{-69}$ |
|  | Score | $1.5 \times 10^{-66}$ | 0 | $4.6 \times 10^{-146}$ |
| Diabetes | SHD | 1 | $1.5 \times 10^{-3}$ | $1.0 \times 10^{-4}$ |
|  | ESHD | 1 | $1.2 \times 10^{-13}$ | $3.1 \times 10^{-5}$ |
|  | Score | $4.8 \times 10^{-1}$ | $2.1 \times 10^{-4}$ | $7.3 \times 10^{-8}$ |
| HailFinder | SHD | 1 | $2.0 \times 10^{-153}$ | $2.6 \times 10^{-62}$ |
|  | ESHD | 1 | $2.5 \times 10^{-169}$ | $2.0 \times 10^{-58}$ |
|  | Score | $9.1 \times 10^{-1}$ | $4.3 \times 10^{-23}$ | $1.3 \times 10^{-143}$ |
| Mildew | SHD | 1 | $1.5 \times 10^{-111}$ | $9.3 \times 10^{-54}$ |
|  | ESHD | 1 | $1.3 \times 10^{-77}$ | $1.9 \times 10^{-49}$ |
|  | Score | 1 | $1.5 \times 10^{-85}$ | $8.3 \times 10^{-79}$ |
| Win95pts | SHD | 1 | $5.7 \times 10^{-209}$ | $5.4 \times 10^{-12}$ |
|  | ESHD | 1 | $8.7 \times 10^{-210}$ | $1.4 \times 10^{-13}$ |
|  | Score | $5.8 \times 10^{-1}$ | $2.6 \times 10^{-26}$ | $3.2 \times 10^{-38}$ |

Table 21: p-values for Conover's squared ranks test, 10,000 samples

|  | Alarm | Barley | HailFinder | Mildew |
|---|---|---|---|---|
| MMHC | $3.2 \times 10^{-2}$ | $2.1 \times 10^{-8}$ | $2.1 \times 10^{-8}$ | $2.1 \times 10^{-8}$ |
| OR1 $k = 5$ | $5.9 \times 10^{-4}$ | $2.1 \times 10^{-8}$ | $2.1 \times 10^{-8}$ | $2.1 \times 10^{-8}$ |
| OR1 $k = 10$ | $1.9 \times 10^{-5}$ | $2.1 \times 10^{-8}$ | $2.1 \times 10^{-8}$ | $2.1 \times 10^{-8}$ |
| OR1 $k = 20$ | $4.1 \times 10^{-8}$ | $2.1 \times 10^{-8}$ | $2.1 \times 10^{-8}$ | $2.1 \times 10^{-8}$ |
| OR2 $k = 5$ | $3.5 \times 10^{-2}$ | $2.1 \times 10^{-8}$ | $2.1 \times 10^{-8}$ | $2.1 \times 10^{-8}$ |
| OR2 $k = 10$ | $1.4 \times 10^{-7}$ | $1.1 \times 10^{-5}$ | $2.1 \times 10^{-8}$ | $2.1 \times 10^{-8}$ |
| OR2 $k = 20$ | $2.1 \times 10^{-8}$ | $3.8 \times 10^{-6}$ | $2.1 \times 10^{-8}$ | $2.1 \times 10^{-8}$ |
| SC $k = 5$ | $2.1 \times 10^{-8}$ | $2.1 \times 10^{-8}$ | $2.1 \times 10^{-8}$ | N/A |
| SC $k = 10$ | $8.1 \times 10^{-1}$ | N/A | N/A | N/A |
| GS | $2.1 \times 10^{-8}$ | $4.3 \times 10^{-7}$ | $2.1 \times 10^{-8}$ | $2.1 \times 10^{-8}$ |
| PC | $4.0 \times 10^{-1}$ | $2.1 \times 10^{-8}$ | $2.1 \times 10^{-8}$ | $2.1 \times 10^{-8}$ |
| TPDA | $6.5 \times 10^{-4}$ | $2.1 \times 10^{-8}$ | $2.1 \times 10^{-8}$ | $2.1 \times 10^{-8}$ |
| GES | N/A | $2.1 \times 10^{-8}$ | $2.1 \times 10^{-8}$ | $2.1 \times 10^{-6}$ |
| Total | $3.3 \times 10^{-2}$ | $1.6 \times 10^{-5}$ | $2.5 \times 10^{-7}$ | $2.3 \times 10^{-6}$ |

Table 22: p-values comparing ACO-E against state-of-the-art algorithms

The results given in Table 13 appear to be indicative of the results as given in Section 8.1. As discussed there, ACO-E had some effect with learning in the Alarm network, especially against a straight greedy search. However, most of this effectiveness appeared to come from the randomness of the search, and did not make much use of the $\rho$ and $\beta$ parameters.

On the networks which ACO-E performed well, Barley and Mildew, this performance is reflected across to the current results as it also performed well here. The results for the HailFinder network seem odd, as the ACO-E algorithm did no better than a search using GREEDY-E. However, in this figure, the performance of GES can also seen to be doing well. As GES works in the space of equivalence classes of Bayesian networks, it is postulated that ACO-E performs well because of the structure of the search space.

As ever, comparisons must be taken tentatively, especially in this case, as the results given by Tsamardinos, Brown et al. only have five samples.

## 8.5 Computational Complexity of ACO-E

The results of experimental Condition 3 show two figures (11 and 12) related to the computational complexity of ACO-E. The first shows that total number of score function evaluations during the algorithm run, the second shows the number of distinct score function evaluations. Both of these are counted, as score function evaluations are usually cached in order to improve running times.

It can be surmised that in general, larger sample sizes imply more evaluations. This makes sense, as larger samples can support networks with more arcs. Since the algorithm starts as the empty graph, it would take more moves and hence more evaluations to get to a maximum. It can also be seen that the total number of evaluations is in general, linear with respect to the number of iterations passed.

Looking at the plots of Figure 12, it can be seen that the number of distinct function evaluations is many magnitudes less than the total number of evaluations. It can also be noted that most of the distinct function evaluations take place within the first twenty to thirty iterations and gradually tails off in a logarithmic fashion. This is to be expected, as at the beginning, the algorithm will explore many new paths when the pheromone is more evenly distributed. The scores for these paths will be cached and so not have to be computed again. This means that over time, less and less new score function applications will be needed. However, it is worthwhile noticing that in many cases the plots do not level out. This implies that new paths are being taken and that the algorithm is not stagnating.

To finish up, it is worthwhile comparing the complexity of ACO-E to the other metaheuristic algorithms that were tested. Looking at Table 17, it appears that ACO-E has a much higher computational complexity than the other algorithms. However, it can be seen that most of these evaluations are not distinct. Since evaluations are normally cached and cache lookup can proceed in constant time, the total score evaluation results are not too important. Focusing instead on distinct score evaluation results, it can be seen that there is not much difference between ACO-E and the other algorithms in terms of actual score function evaluations. Since this is often the dominant factor in algorithm running time, the complexity of the algorithms can be observed as quite similar.

## 9. Conclusions and Future Directions

The main results in this paper were on the development of the ACO-E algorithm as an implementation of the ACO metaheuristic to the problem of learning a Bayesian network structure that provides a good fit to a set of data. In a nutshell, ACO-E performed well in reconstructing test networks, from which data was sampled. A more detailed look at the behavior of ACO-E depending on its parameters, the type of test network and compared to other algorithms will now be given.

## 9.1 ACO-E Behavior as its Parameters are Varied

In analyzing the behavior of ACO-E as a function of its parameters, the best and worst performing figures were compared, across each range of parameter. The best result was found when the parameter setting produced either the highest score or the smallest difference from the test network. The worst result was found when the parameter was 'switched off', i.e. when it had no effect on the algorithm's behavior.

For all parameters, there was a difference between the behavior of the best and the worst settings. Whether this difference was significant or not depended on the particular network being used as a test; some networks responded better to the algorithm than others. For those networks that ACO-E worked well with, the following trends were noticed:

- For data with more features, lower values of $\rho$, higher values of $q_0$ and higher values of $\beta$ worked better; and

- For data with less features, higher values of $\rho$, lower values of $q_0$ and lower values of $\beta$ worked better;

where $\rho$ is the rate of pheromone deposition/evaporation, $q_0$ is the balance between exploration and exploitation and $\beta$ is the power of the heuristic in the probabilistic transition rule.

## 9.2 The Utility of ACO-E as a Function of the Test Network and Sample

It was noticed that ACO-E performed better on some of the test networks than others. The networks that it fared best with were Barley, Mildew and Win95pts, described in Section 6.1.1. On closer examination of these networks it was found that they had a large average v-structure (as discussed in Section 6.1.1) per node value.

The reason that this might make a difference is because nodes with a large number of v-structures imply more possible local maxima in the search space. Greedy methods would run into these maxima, whereas ACO-E is able to find its way around them because of its stochastic nature of not always choosing the best move. Experiments were run to estimate the average v-structure per node value and a correspondence was found in the large sample case. In general, the method used in the experiment could be used to estimate the usefulness of ACO-E in particular situations.

## 9.3 ACO-E Performance Compared to Similar Algorithms

The results of Sections 7.1 and 7.4 show that ACO-E performs well against other algorithms that are similar in nature. These other algorithms were:

- GREEDY-E, which performs a greedy search in the space of equivalence classes of Bayesian network structures (Chickering, 2002a);

- EPQ, which performs an evolutionary programming search in the space of equivalence classes (Cotta & Muruzábal, 2004; Muruzábal & Cotta, 2004); and

- ACO-B, which performs a search using ACO in the space of DAGs (de Campos, Fernández-Luna et al., 2002).

In all cases, the BDeu score of ACO-E was better than the score of the other algorithms, at every iteration. In the case of the structural differences, it was better in all cases, except that of the HailFinder network, where the odd behavior of the scoring function meant better BDeu scores implied worse structural differences. Concurring with the discussion above in 9.2, the networks for which ACO-E performed best were the Barley, Mildew and Win95pts networks.

ACO-E was also shown to be comparable in computational complexity to the other metaheuristic algorithms.

## 9.4 ACO-E Performance Compared to Alternative State-of-the-Art Algorithms

Similar to the section above, ACO-E performed well in comparison to other state-of-the-art Bayesian network structure learning algorithms, performing better in 3 out the 4 tested: Barley, Mildew and HailFinder. The first two are networks in which it performed well in the self test. With the HailFinder network it is postulated that the results are good because of the search space; good results were also shown for the greedy equivalent search (GES) algorithm, which also searches through the space of equivalence classes.

Whilst ACO-E did not perform best with the Alarm network, it did not perform badly either, coming joint third in the rankings. The reasons for the performance on the Alarm network are discussed in Section 8.2.

## 9.5 Extending ACO-E to Increase Performance and Scalability

Since validity checking is the slowest part of the ACO-E algorithm, it currently remains the first issue which must be dealt with, in order to improve running times. However, if that problem is solved then the focus will turn back to the other parts of the algorithm, particularly the scoring function.

**Reducing the Number of Scoring Function Evaluations**  One very easy way in cutting down the number of score evaluations would be to have a static heuristic defined that could say, e.g. what would be the benefit of adding an arc to the empty graph. In this way, scoring functions would only have to be evaluated once per move and hence lead to a speeding up of the algorithm. With a situation like this, local search would become more important in order to 'finish off' traversals to the best possible positions.

**Pruning the Search Space**  Recently, hybrid learning algorithms have shown good success in learning Bayesian network structures, whilst cutting down on running time, sometimes dramatically. They generally work by using a conditional independence test to discover nodes that would likely be connected to a given node and remove the rest of them from consideration. This has the effect of requiring less scoring function evaluations, thus speeding up the algorithm and requiring less memory to store the results of evaluations. With no bound on the number of possible parents, the number of cached values would grow at least quadratically with the number of variables and eventually exhaust the computer's memory.

**Applying ACO-E with Different Search Operators to Better Avoid Local Maxima**  According to Castelo and Kočka (2003), there are certain operators that are able to avoid local maxima in a search space, provided that the sample size tends to infinity. An example of these are the operators given by Chickering (2002b) that are used in a greedy search in the space of equivalence classes of structures (GES).

However, at small sample sizes these guarantees are not strictly true and search algorithms can still get caught in maxima. An example of a method that tries to avoid these is the KES algorithm of Nielsen, Kočka, and Peña (2003), which uses the operators in GES, but has a parameter that controls how often the algorithm acts greedily; when the algorithm does not act greedily, it chooses a move that is not necessarily the best. Experiments show that KES behaves better than GES most of the time.

This procedure bears some similarities to ACO-E. If the randomness was augmented by pheromone and heuristics, there is a possibility that performance would improve even more.

## Acknowledgments

The authors are grateful to the Associate Editor and reviewers for their comments, which were very helpful in guiding the revision of this research.